\documentclass[journal,twoside]{IEEEtran}

\usepackage{xcolor}
\usepackage{tabularx}
\usepackage{booktabs}
\usepackage{makecell}
\usepackage{multirow}
\usepackage{graphicx}
\usepackage{amsmath}
\usepackage{cite}
\usepackage[export]{adjustbox}

\usepackage{tikz}
\usetikzlibrary{spy}
\ifCLASSINFOpdf
\else
\fi
\begin{document}
\title{Fast and Efficient Zero-Learning Image Fusion}

\author{Fayez~Lahoud,~\IEEEmembership{Student~Member,~IEEE,}
        Sabine~S\"usstrunk,~\IEEEmembership{Fellow,~IEEE}% <-this % stops a space
\thanks{The authors are with the School of Computer and Communication Sciences, \'Ecole polytechnique F\'ed\'erale de Lausanne, 1015 Lausanne, Switzerland. E-mail: (fayez.lahoud@epfl.ch;sabine.susstrunk@epfl.ch).}% <-this % stops a space
}

\maketitle

%auto-ignore
\begin{abstract}
We propose a real-time image fusion method using pre-trained neural networks. Our method generates a single image containing features from multiple sources. We first decompose images into a base layer representing large scale intensity variations, and a detail layer containing small scale changes. We use visual saliency to fuse the base layers, and deep feature maps extracted from a pre-trained neural network to fuse the detail layers. We conduct ablation studies to analyze our method's parameters such as decomposition filters, weight construction methods, and network depth and architecture. Then, we validate its effectiveness and speed on thermal, medical, and multi-focus fusion. We also apply it to multiple image inputs such as multi-exposure sequences. The experimental results demonstrate that our technique achieves state-of-the-art performance in visual quality, objective assessment, and runtime efficiency.	
\let\thefootnote\relax\footnotetext{Code available at https://github.com/IVRL/Fast-Zero-Learning-Fusion}
\end{abstract}

\begin{IEEEkeywords}
Image fusion, visual saliency, two-scale decomposition, neural networks, real-time
\end{IEEEkeywords}

% \fi
\IEEEpeerreviewmaketitle

%auto-ignore
\section{Introduction}

\IEEEPARstart{I}{mage} fusion plays an important role in multiple image processing and computer vision applications. In fact, any procedure requiring the analysis of two or more images of the same scene benefits from image fusion~\cite{li2017pixel}. For instance, image fusion between visible and infrared bands~\cite{ma2019infrared} is used in night-time surveillance, image dehazing~\cite{dumbgen2018near}, face recognition~\cite{ma2016infrared}, military reconnaissance missions, and firefighting~\cite{lahoud2018ar}. In medical imaging applications, images can come from multiple modalities such as magnetic resonance imaging (MRI), positron emission tomography (PET), and computed tomography (CT). The fusion of these images helps physicians provide reliable and accurate medical diagnosis to their patients, and navigate otherwise impossible surgeries~\cite{du2016overview}. In photography, the fusion of images taken with different focal settings allows photographers to obtain a single all-in-focus image~\cite{garg2014survey,el2017correlation}. Finally, a stack of low dynamic range images with varying exposure levels can be fused into a single high dynamic range image~\cite{ma2017robust,ram2017deepfuse} that usually looks better.

A large number of image fusion techniques have been proposed in the literature~\cite{garg2014survey,du2016overview,li2017pixel,ma2019infrared}. Classical approaches often rely on multi-scale fusion techniques~\cite{toet1989morphological,singh2014fusion}. These methods excel at preserving details from different source images. However, they may produce brightness and color distortions since they do not consider spatial consistency in their fusion process. Other approaches use sparse representation~\cite{liu2015general,liu2016image}, or pulse-coupled neural networks~\cite{wang2013multimodal,yin2018medical}, and rely on optimization techniques  with multiple iterations. Although they can produce better fusion results, they are computationally inefficient. Finally, all these methods rely on manually crafted fusion rules that might not apply for all source images.

Recently, convolutional neural networks (CNN) have been successfully used to design robust and efficient fusion rules for multiple fusion tasks\cite{liu2017multi,ram2017deepfuse,li2018infrared}, with state-of-the-art performance. In fact, neural networks can be considered as feature extractors, where intermediate maps represent salient features that can be used to generate fusion weight maps. Unlike classical methods, which are based on a single fusion rule, CNNs are trained on large datasets and can model a variety of image features, allowing them to obtain more general fusion rules. However, training these networks requires the construction of large datasets, which is expensive and time-consuming. They are also specialized to single tasks and do not perform as well on tasks they were not trained for. Finally, they require large amounts of memory, time, and energy to run.

\begin{figure*}
	\input{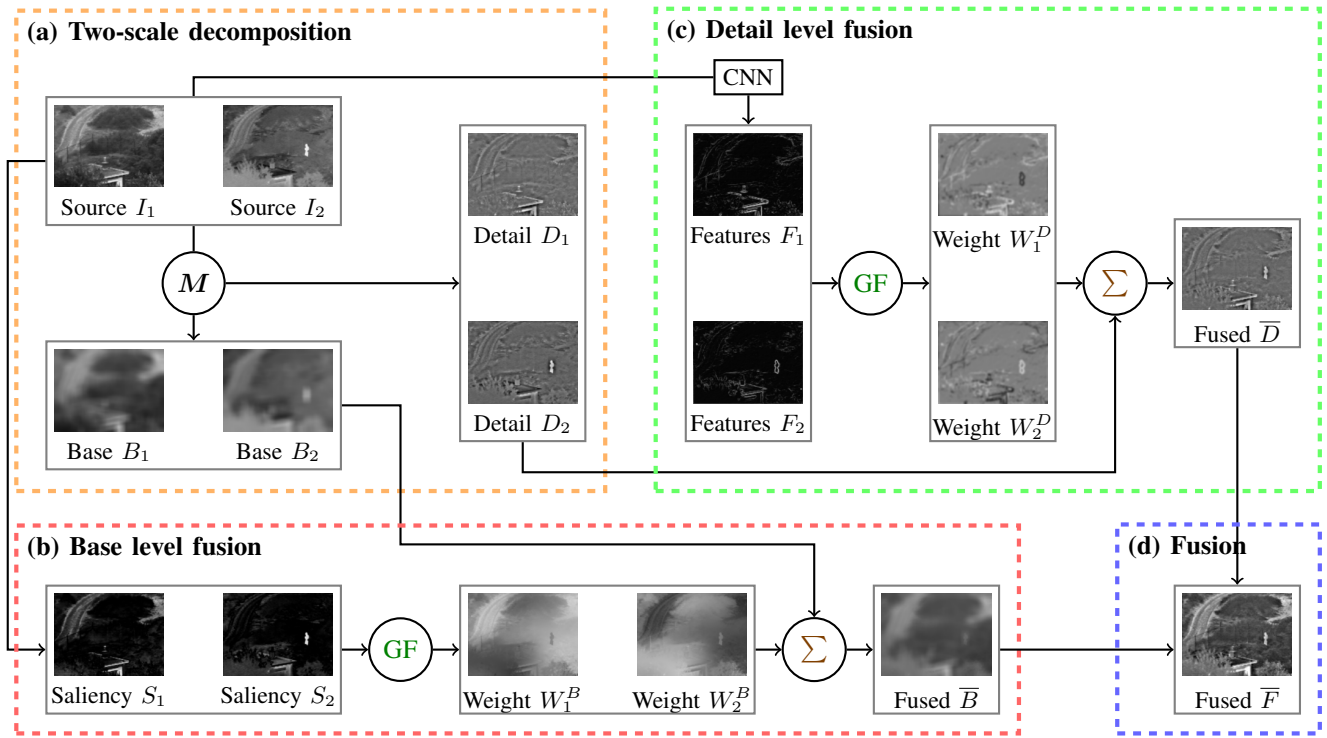}
	\caption{\label{fig:full-diagram} Schematic diagram of our proposed fusion method. The two-scale decomposition transforms the image into base and detail layers. The base layers are fused based on a saliency measure comparison, and the detail layers based on CNN intermediate feature maps. The guided filter is used to smooth the weight maps and enforce consistency with the source images. In the last stage, the fused base and detail layers are joined to obtain the final fusion. $M$, \textcolor{green!50!black}{GF}, and \textcolor{orange!50!black}{$\sum$} indicate the two-scale decomposition filter, the input-guided filter, and the weighted sum operator, respectively.}
\end{figure*}

To solve the aforementioned	problems, we propose a novel and fast image fusion based on \textit{pre-trained} convolutional neural networks for image fusion, illustrated in Fig.~\ref{fig:full-diagram}. We first decompose the source images into base and detail layers. We use visual saliency to fuse the base layers, and deep feature maps from a pre-trained CNN to fuse the detail layers. The weights maps are aligned with the source images through a guided filter. Then, both fused base and detail levels are joined to reconstruct the fused image. 

Our approach is based on pre-trained neural networks, which have learned a multitude of example images and incorporated a large set of image priors. This, contrasted with MST-, SR-, and PCNN-based methods, allows our technique to generalize better on various images and image fusion tasks. Additionally, unlike prior CNN-based fusion techniques, our method requires neither a specific dataset nor a particular network architecture since it leverages pre-trained network models. Therefore, our method alleviates the need to collect training data, and generalizes well to any fusion problem.

We conduct experiments to analyze the proposed method and its parameters. First, we study the effect of the decomposition filter on the fusion quality and its runtime. Second, we run ablation studies to understand the impact of the weight construction techniques, the guided filter parameters, and the depth and architecture of the pre-trained CNNs on the fusion performance. The ablation studies show the advantages of our fusion rules and their robustness to changes in parametrization. We compare our approach with state-of-the-art methods on thermal, medical, and multi-focus fusion. Our technique demonstrates advantages in visual quality, objective assessment, and runtime efficiency. Finally, we show that our method is applicable to any number of source images by displaying multi-exposure sequence fusion results.

Our main contribution is a novel technique to integrate pre-trained neural networks in an image fusion pipeline. Our method shows state-of-the-art performance and runs in real time. The low computational costs make it very beneficial for embedded systems applications, such as thermal fusion for firefighters where security restrictions enforce hardware limitations, medical fusion for continuous monitoring systems and efficient diagnosis, or multi-focus and multi-exposure fusion integrated in camera pipelines.

%auto-ignore
\section{Related Work}
We present a two-scale image fusion method based on saliency and pre-trained convolutional neural networks. Consequently, we review classical and CNN-based image fusion approaches. We discuss two-scale decompositions, visual saliency, and edge-preserving filters in the context of image fusion. Finally, we address pre-trained CNNs and their ability to represent images in a space well-suited for comparison.

\subsection{Image fusion}
Classical fusion algorithms use multi-scale transforms (MST) to obtain perceptually good results. Frequently used MSTs include pyramids~\cite{toet1989morphological,du2016union}, wavelets~\cite{qu2001medical,singh2014fusion}, filters~\cite{li2013image,kumar2015image}, and geometrical transforms like contourlets and shearlets~\cite{das2012nsct,bhatnagar2013directive,wang2013multimodal}. Another popular approach for modeling images is sparse representation (SR)~\cite{zhang2013dictionary,liu2015general,liu2016image,liu2017infrared}. Under SR, images are encoded as a sparse linear combination of samples representing salient features. This encoding is used to fuse the source images. Pulse-coupled neural networks (PCNN) are another set of models that are commonly used in image fusion~\cite{wang2013multimodal,yin2018medical}. They are inspired by the cat's visual cortex, and are two-dimensional, with each neuron corresponding to a single pixel in an image. MST, SR, and PCNN approaches enforce a specific relation between inputs and outputs, i.e., they operate on a rule which fits a limited set of predefined image priors. Our approach is based on pre-trained CNNs which learn by example. Having been trained on a large amount of examples, these networks are able to construct a more complete understanding of image priors, and as such generalize better on image fusion tasks.

The earliest fusion work involving neural networks poses multi-focus fusion as a classification task~\cite{li2002multifocus}. Three focus measures define the input features to a shallow network which outputs the weight maps corresponding to the source images. Due to architectural constraints, the method can only run on image patches, and generates boundary artifacts. More recently, convolutional neural networks have been trained to generate decision maps for multi-focus~\cite{liu2017multi}, multi-exposure~\cite{ram2017deepfuse}, medical~\cite{liu2017medical}, and thermal fusion~\cite{li2018infrared}. Although these approaches often achieve better performance than their classical counterparts, they still have major drawbacks. First, they require large specialized datasets for training. Second, deep networks often overfit the datasets they are trained on, e.g., a network trained for multi-focus image fusion will only be suitable for that task. Finally, CNNs require large amounts of memory and are computationally expensive both in time and energy. In contrast, our method requires no training, which alleviates the necessity of collecting data. Moreover, we use a single pre-trained network as feature extractor for any image fusion task, so our technique generalizes well to any fusion problem. Finally, by evaluating the impact of depth and architecture, we can select a memory and time efficient neural network for our fusion method.

\subsection{Two-scale decomposition}
Two-scale decompositions are a subset of MSTs that are commonly used for image enhancement and tone mapping~\cite{farbman2008edge,subr2009edge,gastal2011domain}. The two-scale approach avoids mixing low and high frequency information and reduces halo artifacts. It operates in the spatial domain, and preserves source intensity values and spatial constraints. The base layer contains large scale intensity variations and is obtained by applying a smoothing filter on the image. The detail layer is the difference between the original image and the base layer and contains small scale information. Multiple decomposition filters have been proposed and used for image fusion such as the average filter~\cite{li2013image,bavirisetti2016two}, bilateral filter~\cite{tomasi1998bilateral,hu2012multiscale,kumar2015image}, and local extrema~\cite{subr2009edge,xu2014medical}. We conduct ablation studies to choose the most appropriate filter for our method.

\subsection{Visual saliency}
Visual saliency detects perceptually salient visual structures, regions or objects in an image~\cite{zhai2006visual,achanta2009frequency,cheng2015global}. These locations stand out from their neighborhood, and attract the human visual system's attention. Saliency is widely used in multiple computer vision tasks such as object recognition~\cite{wu2013scale} and person re-identification~\cite{zhao2013unsupervised}. As for image fusion, saliency maps commonly serve as activity level measurements, since they reflect the important features of the source images~\cite{liu2017infrared,ma2017infrared}. We use saliency to fuse the base layers as it correlates with the contrast between different low-frequency image regions. We use the method presented in~\cite{zhai2006visual} because it can produce saliency maps in $O(N)$ time with low memory requirements.

\subsection{Edge-preserving filters}
Edge-preserving filters such as the bilateral filter~\cite{tomasi1998bilateral}, guided filter~\cite{he2010guided}, and cross-bilateral filter~\cite{petschnigg2004digital} smooth image details while preserving strong edges and avoiding ringing artifacts. Due to these qualities, they are used in image fusion either as multi-level decomposition methods~\cite{hu2012multiscale,kumar2015image,li2015weighted} or consistency verification schemes~\cite{li2013image,liu2017multi}. Among these filters, the guided filter~\cite{he2010guided} is a time-efficient method that runs in linear time, independently of the filter size. It is used in multiple applications such as detail enhancement and no-flash denoising~\cite{he2010guided}, and in image fusion methods~\cite{li2013image,li2015weighted}. We use it as a consistency verification model to ensure that the generated weight maps adhere to the region boundaries defined by the source images.

\subsection{Pre-trained models}
Pre-trained neural networks are networks that have been previously trained on a specific task and dataset. While these networks are specifically built for that task in mind, they can be used to generate image representations for other tasks. A simple example would be to use the neural networks as feature extractors instead of manually designing shape, color, and texture features for classification~\cite{sermanet2013overfeat}. Additionally, neural style transfer compares deep features from a pre-trained network to transfer style statistics from a target image to a source image~\cite{gatys2016image}. Deep features can also be stacked together to form hypercolumns for semantic image segmentation~\cite{hariharan2015hypercolumns}. These works show that deep features are well-suited to compare texture, style, and semantics between images without any additional training or data collecting costs. Similarly, we use pre-trained neural networks to generate detail fusion weights.

%auto-ignore
\section{CNN-based image fusion}\label{sec:method}
The schematic in Fig.~\ref{fig:full-diagram} summarizes the main sections of the proposed two-scale decomposition fusion method. First, an image filter is applied to divide the image into base and detail levels. Then the base and detail layers are fused based on saliency measures and neural network activations, respectively. At the end, the fused representations are joined together to reconstruct the final fused image.

\subsection{Two-scale decomposition}
Suppose there are $K$ pre-registered source images denoted as $I_k | k \in \{1,2,\dots,K\}$. As seen in Fig.~\ref{fig:full-diagram}(a), each source images $I_k$ is decomposed into two-scale levels using a smoothing image filter. The base layer of source image $I_k$ is the result of the smoothing operation using filter $M$
\begin{equation}
B_k = M(I_k) .
\end{equation}
Given the base layer $B_k$, the detail layer is obtained by subtracting it from the source image $I_k$ as
\begin{equation}
D_k = I_k - B_k .
\end{equation}

This two-level decomposition separates the image into two components, the first containing large-scale intensity variations and smooth regions and another containing small-scale edges and image details such as textures. The choice of smoothing filter and its effect on the quality and runtime of the fusion are discussed in the experimental evaluations.

\subsection{Base layer fusion}
In order to fuse the base layers $B_k | k \in {1, 2,\dots,K}$, we first compute the visual saliency map of each image $I_k$, as illustrated in Fig.~\ref{fig:full-diagram}(b). Let $I(p)$ denote the intensity value of pixel $p$ in a given image $I$. The saliency value $S(p)$ at pixel $p$ is defined as
\begin{equation}
\begin{split}
S(p) & = |I(p) - I(1)| + |I(p) - I(2)| + \dots + |I(p) - I(N)| \\
& = \sum_{i=0}^{255} M(i) |I(p) - i| ,
\end{split}
\end{equation}
where $N$ is the total number of pixels in the image, and $M(i)$ is the frequency of pixels whose value is $i$, i.e., the histogram of the image at $i$. The histogram can be computed in $O(N)$ time order, and the distance measures $|I(p) - i|$ can be calculated for all $i \in [0, 255]$ in constant time. By pre-computing all color distances, a map $D(x, y) = |I(x) - I(y)|$ can be constructed for all color value pairs. Given the histogram $M(.)$ and the color distance matrix $D(., .)$, the saliency for pixel $p$ is simply computed as
\begin{equation}
S(p) = \sum_{i=0}^{255}M(i)D(p,i) ,
\end{equation}
which also completes in constant time. The saliency maps are then normalized to the range $[0, 1]$. For further details, please refer to the complete implementation in~\cite{zhai2006visual}.

Following the computation of saliency maps $S_k$ for each image $I_k$, we generate the base layer weight maps $W_k^b$ as follows
\begin{equation}
W_k^b = \frac{S_k}{\sum_{j\in K} S_j} .
\end{equation}

The weight maps $W_k^b$ obtained are generally noisy and do not completely respect region boundaries, which could further produce artifacts in the fused image. Enforcing spatial consistency via guided filtering solves this problem: the filtering process ensures that adjacent pixels with similar intensity values obtain similar weights. This leads to smooth weight maps for the base layers without introducing artificial edges. The smoothed weights $\overline{W_k^b}$ are thus obtained as
\begin{equation}
\overline{W_k^b} = G_{r_b,\epsilon_b}(W_k^b, I_k) ,
\end{equation}
where $G_{r,\epsilon}(., .)$ is the guided filter, and $r$ and $\epsilon$ control the radius and degree of blur, respectively. After obtaining the $K$ weight maps, they are all normalized such that they sum up to one at each pixel $p$
\begin{equation}\label{eq:normalization}
\overline{W_k^b} = \frac{\overline{W_k^b}}{\sum_{j\in K} \overline{W_j^b}} \ .
\end{equation}

Following that, the fused base layer is constructed as a weighted-sum of all base layers given the weight maps
\begin{equation}
\overline{B} = \sum_{k \in K} \overline{W_k^b} B_k .
\end{equation}

\subsection{Detail layer fusion}
We propose a novel fusion strategy using convolutional neural networks to extract deep features and generate weight maps. In this section, we detail how we obtain the weight maps, and construct the fused image. Fig.~\ref{fig:full-diagram}(c) illustrates the process of taking two source images, extracting their representative feature maps, generating smoothed fusion weights according to their activation levels, and finally using them to fuse the detail layers.

Recall that there are $K$ pre-registered source images denoted as $I_k | k \in \{1, 2, \cdots, K\}$. Additionally, suppose there is a pre-trained convolutional neural network with $\mathcal{L}$ layers, with $C_l$ output channels per layer $l$. We denote $F_k^{c, l}$ as the $c$-th feature map of the $k$-th image extracted at the $l$-th layer of the network (taken after ReLU operation), the feature map is computed as
\begin{equation}
F_k^l = \max(0, \mathcal{F}_l(I_k)),
\end{equation}
where $\mathcal{F}_l(.)$ is the application of the network function to the input image up to layer $l$. The $\max(0, .)$ function denotes the ReLU operation.

For every feature map, we denote $\hat{F}_k^l$ as the $l_1$-norm computed over the $C_l$ channels of the feature maps of layer $l$ as follows 
\begin{equation}
\hat{F}_k^l = \sum_{c=0}^{C_l} || F_k^{c,l} ||_1 \ .
\end{equation}
This constitutes a measure of the activity corresponding to the input image at layer $l$ where high pixel values correspond to high activity levels.

The feature maps are extracted for $\mathcal{L}$ layers , so we obtain, per image $k$, a set of features maps $\{\hat{F}_k^l | l \in \mathcal{L}\}$. For every layer $l$, the $K$ feature maps are used to generate $K$ weight maps that indicate the amount of contribution of each image at a specific pixel. For our method, we use the softmax operator to generate said maps as follows
\begin{equation}
W_k^{d,l} = \frac{e^{\hat{F}_k^l}} {\sum_{j=1}^{K} e^{\hat{F}_j^l}} \ ,
\end{equation}
where $e^{(.)}$ is the exponentiation with base $e$.

In order to account for small mis-registrations, and remove undesirable artifacts around the edges of both modalities, we apply a guided filter to the weight maps. This correctly aligns the weights with the source edges while preserving the sharpness of their details. After smoothing the weight maps using the guided filter as
\begin{equation}
\overline{W_k^{d,l}} = G_{r_d,\epsilon_d}(W_k^{d,l}, I_k) ,
\end{equation}
the map are normalized such that they sum up to one at each pixel $p$, similar to Eq.~\eqref{eq:normalization}.

At layer $l$, we have a set of weights $\overline{W_k^{d,l}} | k \in \{1, 2, \cdots, K\}$. Using these weight maps, the image fusion at layer $l$ is computed as
\begin{equation}\label{eq:fuse}
\overline{D^l} = \sum_{k=1}^K \overline{W_k^{d,l}} D_k .
\end{equation}

Finally, some neural networks contain convolution layers with large strides or pooling layers, and thus generate deep feature maps with smaller spatial resolution than the input images. In these cases, we simply upsample the feature maps using nearest neighbors. The subsequent filtering step ensures the weight maps remain smooth and adhere to the edges of the source images even after upsampling.

\subsection{Two-scale reconstruction}
Having obtained both fused base $\overline{B}$ and detail $\overline{D}$ layers, the fused image $\overline{F}$ is the pixel by pixel combination of those layers, as illustrated in Fig.~\ref{fig:full-diagram}(d):
\begin{equation}
\overline{F} = \overline{B} + \overline{D}.
\end{equation}

$\overline{F}$ is clipped to the appropriate range to remove any out of range values. Finally, if desired, a tone mapping function could be applied to the result.

%auto-ignore
\begin{figure}[t]
	\newcommand{\dwi}{0.07\textwidth}
	\newcommand{\dhi}{0.07\textwidth}
	\newcommand{\figone}[1]{
		\includegraphics[width=\dwi,height=\dhi]{#1}
	}

	\begin{center}
		\small
		\newcolumntype{T}{ >{\centering\arraybackslash} m{0.075\textwidth} }
		\begin{tabular}{@{} *{6}{@{\hskip 0.001\linewidth} T}}
			\figone{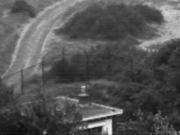} & \figone{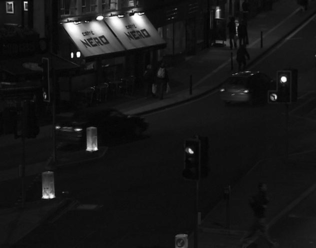} & \figone{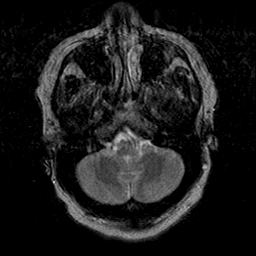} & \figone{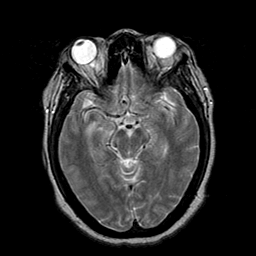} &  \figone{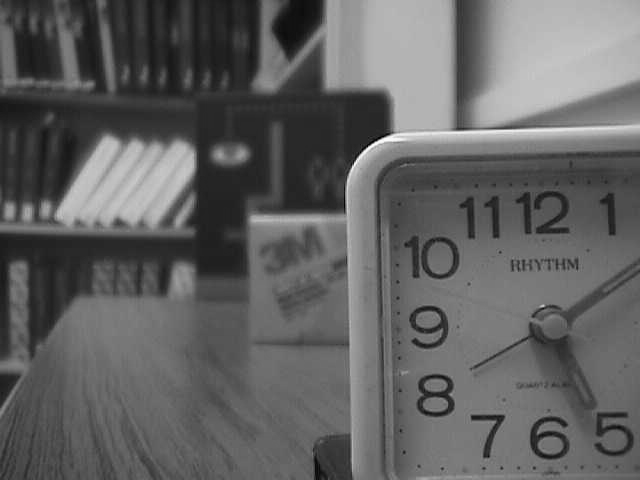} & \figone{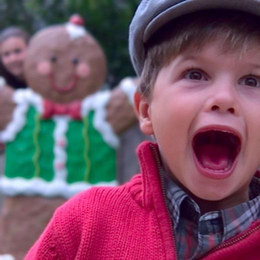} \\
			
			\figone{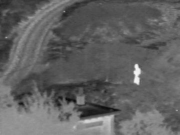} & \figone{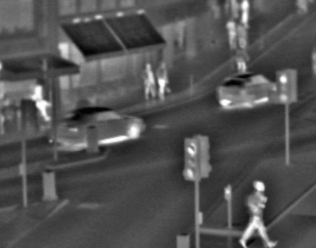} & \figone{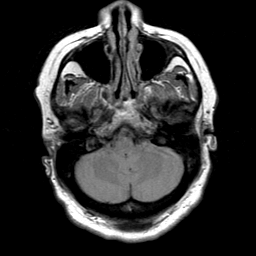} & \figone{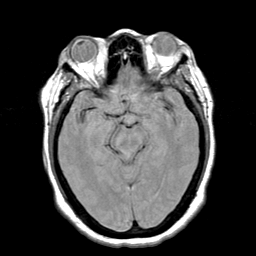} & \figone{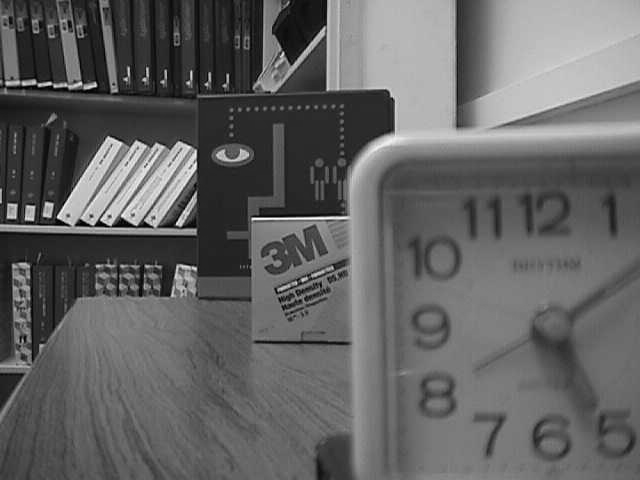} & \figone{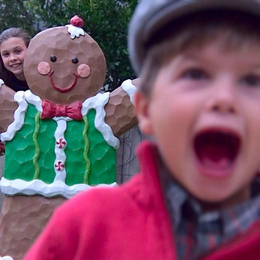} \\
			
			TNO & TNO & MRI-CT & T1-T2 & Focus & Lytro\\
			
			\noalign{\vspace{-0.2cm}}
		\end{tabular}
	\end{center}

	\caption{\label{fig:data-examples} Pairs of testing images from the datasets used in the experiments. From left to right, shown are two pairs of visible and infrared images, two pairs of MRI-T1 and MRI-T2 images, and two pairs of multi-focus images.}
\end{figure}

\section{Experiments}
\subsection{Experimental setup}
We perform evaluations on three fusion tasks using different datasets. The first dataset is the TNO set containing $21$ pairs of visible and infrared images representing natural scenes~\cite{toet2014tno}. The second dataset is extracted from the Whole Brain Atlas~\cite{summers2003harvard}, and contains $97$ pairs of computed tomography (CT) and magnetic resonance imaging (MRI) images, and $24$ pairs of T1-T2 weighted MRI images. CT is sensitive to dense structures and bones while MRI captures softer organ tissues. T1 and T2 are relaxation times that characterize tissue, and thus capture different properties of the same substance under MRI. The third dataset consists of $20$ commonly used images in multi-focus fusion ('Book', 'Clock', 'Desk', etc.) in addition to the multi-focus color images from the Lytro dataset~\cite{nejati2015multi}. The images from these datasets have been used in many related papers: TNO~\cite{zhang2013dictionary,ma2016infrared,ma2017infrared}, Whole Brain Atlas~\cite{li2013image,bhatnagar2013directive,liu2017medical}, multi-focus images~\cite{liu2015multi,liu2017multi,zhang2017boundary}, and Lytro~\cite{li2017pixel,liu2017multi}. All the images are pre-registered. Fig.~\ref{fig:data-examples} shows sample pairs taken from these datasets.

In the following experiments, most of the source images are grayscale and the fusion is as presented in Sec.~\ref{sec:method}. However, for the Lytro dataset, color images are fused via their luma components. The sources are transformed from RGB color space to YCbCr. Then the Y components are fused together while the Cb and Cr components are averaged. The final fused image is reconstructed by converting the intermediate YCbCr fused image back to RGB.

\subsection{Image fusion metrics}
In order to quantitatively evaluate the performance of different fusion methods, we use eight commonly adopted objective fusion metrics. They are entropy EN, mutual information MI, visual information fidelity of fusion VIFF~\cite{han2013new}, mutual information index $Q_{MI}$~\cite{hossny2008comments}, edge information index $Q_G$~\cite{xydeas2000objective}, structural similarity index $Q_Y$~\cite{yang2008novel}, Cvejic's metric $Q_C$~\cite{cvejic2005similarity}, and phase congruence index $Q_P$~\cite{zhao2007performance}. EN reflects the amount of information present in the fused image while MI estimates the amount of information transferred from the inputs images to the fused image. VIFF highly correlates with the human visual system. $Q_{MI}$ measures how well information from every source image is preserved. $Q_G$ measures the amount of edge information transferred. $Q_Y$ measures how well structural information is preserved in the fusion process. $Q_C$ evaluates the method's ability in transferring information while reducing distortions. Finally, $Q_P$ reflects how well salient features are preserved. For all these metrics, a higher value suggests a better performance. Values for VIFF, $Q_G$, $Q_Y$, $Q_C$, and $Q_P$ are normalized values and bounded between $[0, 1]$. EN, MI, and $Q_{MI}$ are bounded by the number of bits used to represent pixels based on the Shannon Entropy rule, e.g., EN is bounded between $[0, 8]$ for images using 8-bit pixel representation $[0-255]$.

%auto-ignore
\subsection{Ablation studies}\label{sec:ablation}
In this subsection, we study the influence of different parameters on our fusion model. First, we conduct a study on two-scale decomposition filters, and compare their impact on the quality and runtime of our method. Then, we compare our weight generation techniques using saliency and CNNs with the max and average fusion rules commonly found in the literature~\cite{bhatnagar2013directive,ma2017infrared,li2018infrared}. As we are using a parameterizable guided filter function, we also analyze the effect of its free parameters on both base and detail layer fusion. Finally, we study the effect of network architecture and layer depth on the fusion in terms of performance, runtime, and memory consumption. The ablation studies are performed on the TNO dataset. Here, we only present plots for $Q_G$, $Q_Y$, $Q_C$, and $Q_Y$. Please refer to the supplemental material for additional results.

\begin{figure}[t]
	\centering
	\includegraphics[width=\linewidth]{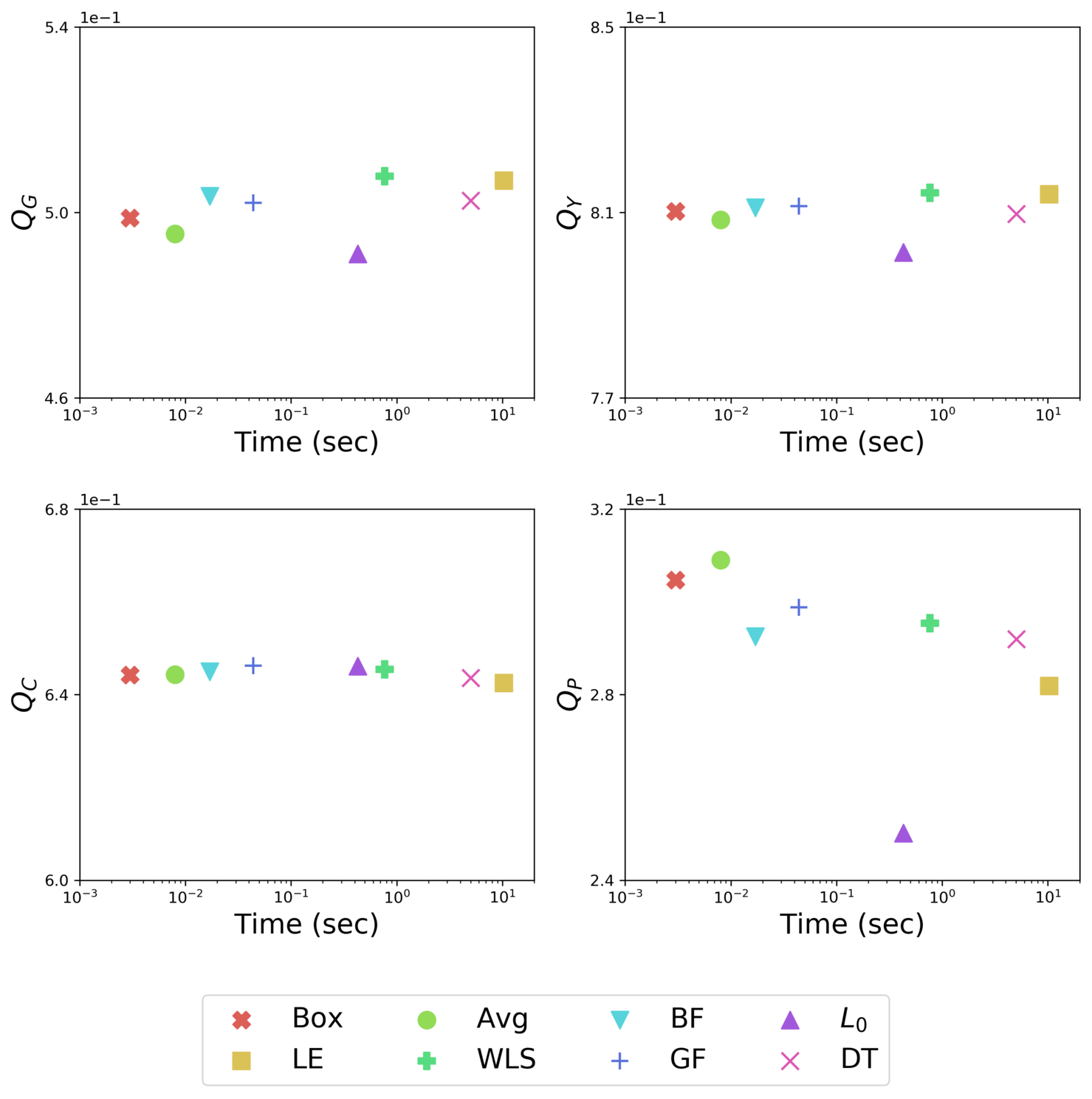}
	\caption{\label{fig:decomp-speed} Fusion quality with respect to two-scale decomposition runtime.}
\end{figure}

\subsubsection{Two-scale decompositions}
In this experiment, we compare commonly used two-scale decomposition filters for their impact on the proposed method. The compared filters are box (Box), average (Avg), bilateral (BF)~\cite{tomasi1998bilateral}, $L_0$-minimization ($L_0$)~\cite{xu2011image}, local extrema (LE)~\cite{subr2009edge}, weighted least squares (WLS)~\cite{farbman2008edge}, guided filter (GF)~\cite{he2010guided}, and domain transform (DT)~\cite{gastal2011domain}. Fig.~\ref{fig:decomp-speed} shows the fusion quality with respect to the two-scale decomposition runtime of each evaluated filter. The quality of the resulting fusion is not heavily affected by the choice of filter, with the maximum difference across all metrics being $0.062$ for $Q_P$ between the $L_0$ and Avg filters on a scale from $[0, 1]$. However, the runtime of these decompositions varies between $0.001$ seconds for the box filter and $10.312$ seconds for the local extrema filter. The plots demonstrate that a simple filter for decomposition is as good as the more complicated edge-aware or minimization-based filters while being orders of magnitude faster.

\subsubsection{Weight construction methods}
\newcolumntype{L}{>{\centering\arraybackslash}m{1.4cm}}
\begin{table}[t]
	\begin{center}
		\caption{\label{tbl:weight-methods} Quantitative assessment of different weighting schemes for both base and detail layers.}
		\begin{tabular}{l *{4}{L}}
			Metric & AVG-MAX & AVG-CNN & S-MAX & S-CNN \\ 
			\noalign{\smallskip} \hline\noalign{\smallskip}
			\textit{EN} & 6.639 & 6.618 & 7.016 & \textbf{7.126} \\
			\textit{MI} & 13.277 & 13.236 & 14.032 & \textbf{14.252} \\
			\textit{VIFF} & 0.562 & 0.625 & 0.610 & \textbf{0.690} \\
			\textit{$Q_{MI}$} & 2.006 & 1.997 & 2.063 & \textbf{2.064} \\
			\textit{$Q_G$} & 0.444 & 0.498 & 0.440 & \textbf{0.500} \\
			\textit{$Q_Y$} & 0.768 & 0.807 & 0.775 & \textbf{0.816} \\
			\textit{$Q_C$} & 0.614 & 0.623 & 0.600 & \textbf{0.649} \\
			\textit{$Q_P$} & 0.251 & 0.294 & 0.270 & \textbf{0.315} \\
			\noalign{\smallskip}\hline
		\end{tabular}
	\end{center}
\end{table}

\begin{figure}[t]
	\centering
	\includegraphics[width=\linewidth]{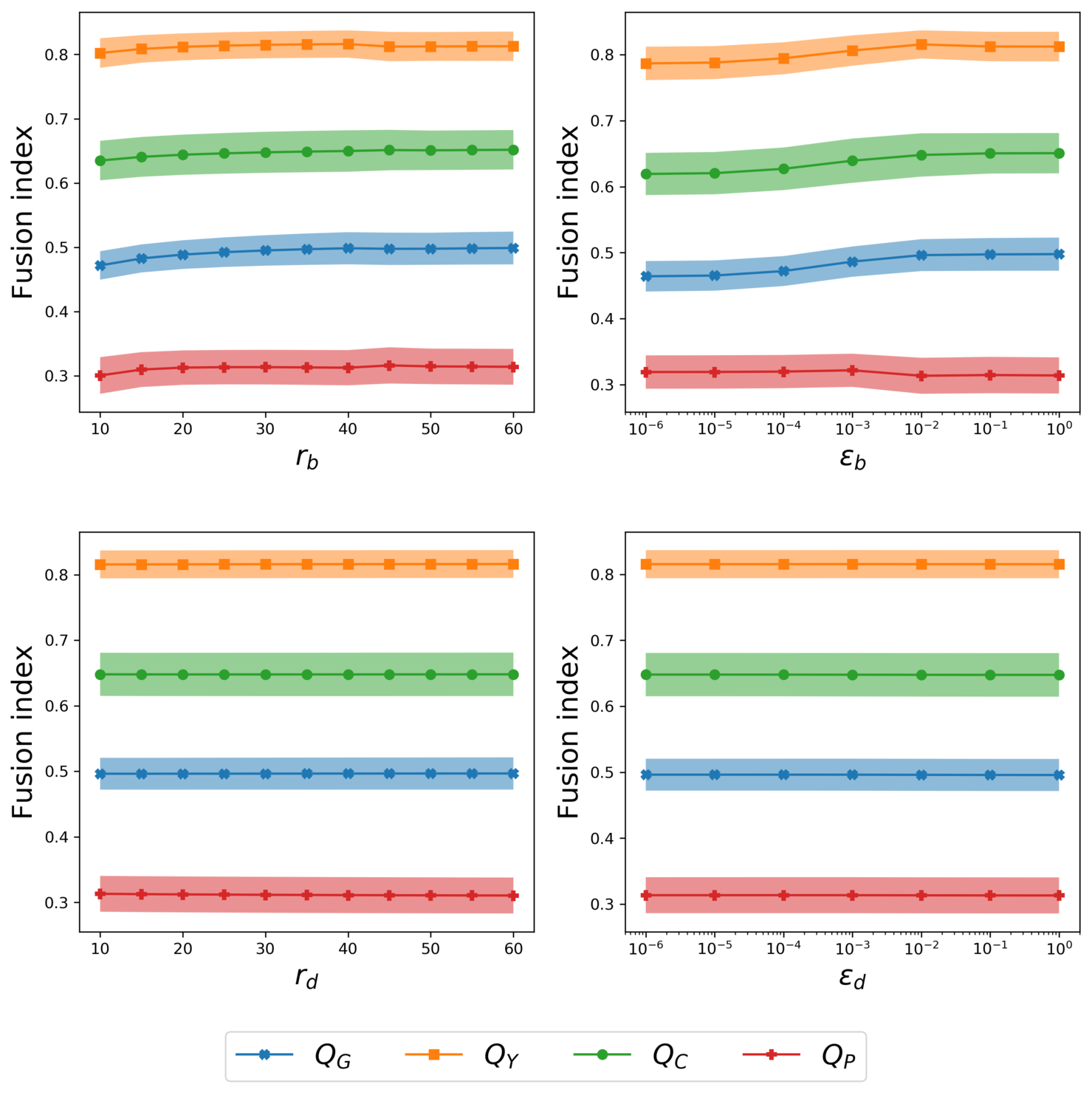}
	\caption{\label{fig:guided-params} Fusion quality with respect to the guided filter parameters. The shaded regions indicate the range of values obtained on the evaluated images.}
\end{figure}

We compare our fusion rules to the commonly used weight construction approaches of averaging (AVG) for base layers, and maximum for detail layers (MAX). In comparison, our fusion rules are saliency (S) for base layers, and (CNN) for detail layers. We compare the following weight construction approaches: AVG-MAX, AVG-CNN, S-MAX, and S-CNN. Table~\ref{tbl:weight-methods} shows the average values of the fusion metrics on the TNO dataset. EN, MI, VIFF, and $Q_P$ significantly improve between AVG-MAX and S-MAX, showing the impact of saliency based fusion on the base layers. Additionally, $Q_G$, $Q_Y$, and $Q_C$ show small differences between AVG-CNN and S-CNN and large differences against their AVG-MAX and S-MAX counterparts, confirming the benefits of using pre-trained deep network features to fuse the detail layers. Together, the saliency and CNN-based fusion weights generate the highest quality fusion results on all evaluated metrics, validating our choices for fusion rules.

\subsubsection{Guided filter parameters}

We use two different guided filters $G_{r_b,\epsilon_b}(., .)$ and $G_{r_d,\epsilon_d}(., .)$, and as such have four different parameters to tune. In order to study the impact of each parameter, we freeze the remaining ones. The base guided filter is frozen to a large size and blur degree to ensure smooth base weight maps. In contrast, the detail guided filter is frozen to a small size and blur degree to preserve sharpness in the detail weight maps. As such, when a parameter is frozen it is set to one of $r_b=35$, $\epsilon_b=0.01$, $r_d=7$, and $\epsilon_d=10^{-6}$. A similar approach was also employed in~\cite{li2013image}. The filter size $r$ is changed linearly from $10$ to $60$ in steps of $5$, and the blur amount $\epsilon$ is changed logarithmically from $10^{-6}$ to $1$ in steps of $10$.

Fig.~\ref{fig:guided-params} shows the changes in objective fusion quality with respect to changes in the guided filter parameters. The plots for $r_b$ and $\epsilon_b$ show that larger values are preferred to obtain a higher quality base layer fusion. However, the choice of $r_d$ and $\epsilon_d$ do not significantly affect the performance, even though higher values lead to slightly lower quality. 

\begin{figure}[t]
	\centering
	\includegraphics[width=\linewidth]{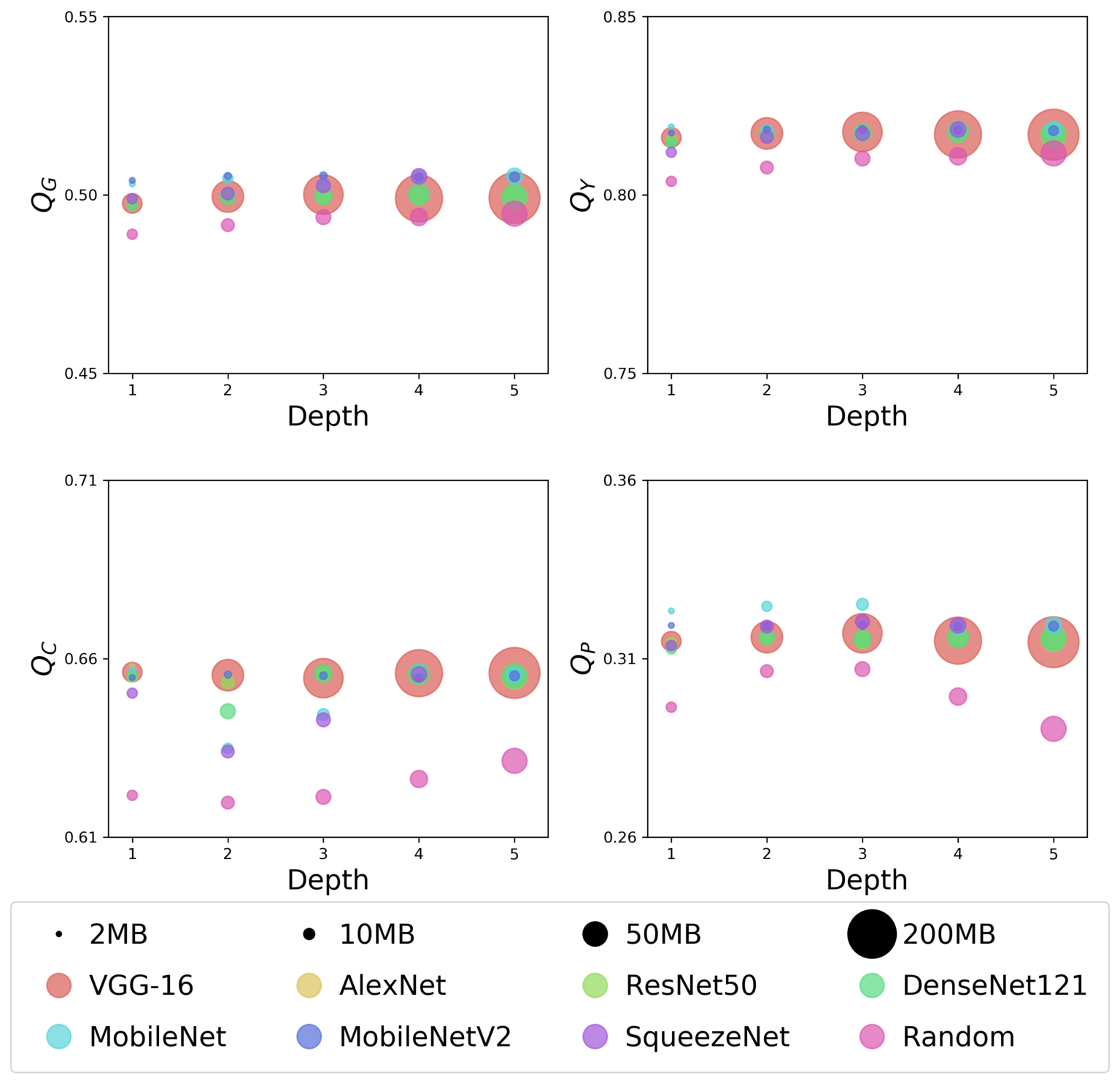}
	\caption{\label{fig:net-depth} Fusion performance with respect to network architecture and depth. The depth indicates the convolutional block at which the deep feature maps are compared. The disk size indicates the memory requirements for the evaluation up-to the specified depth.}
\end{figure}

%auto-ignore
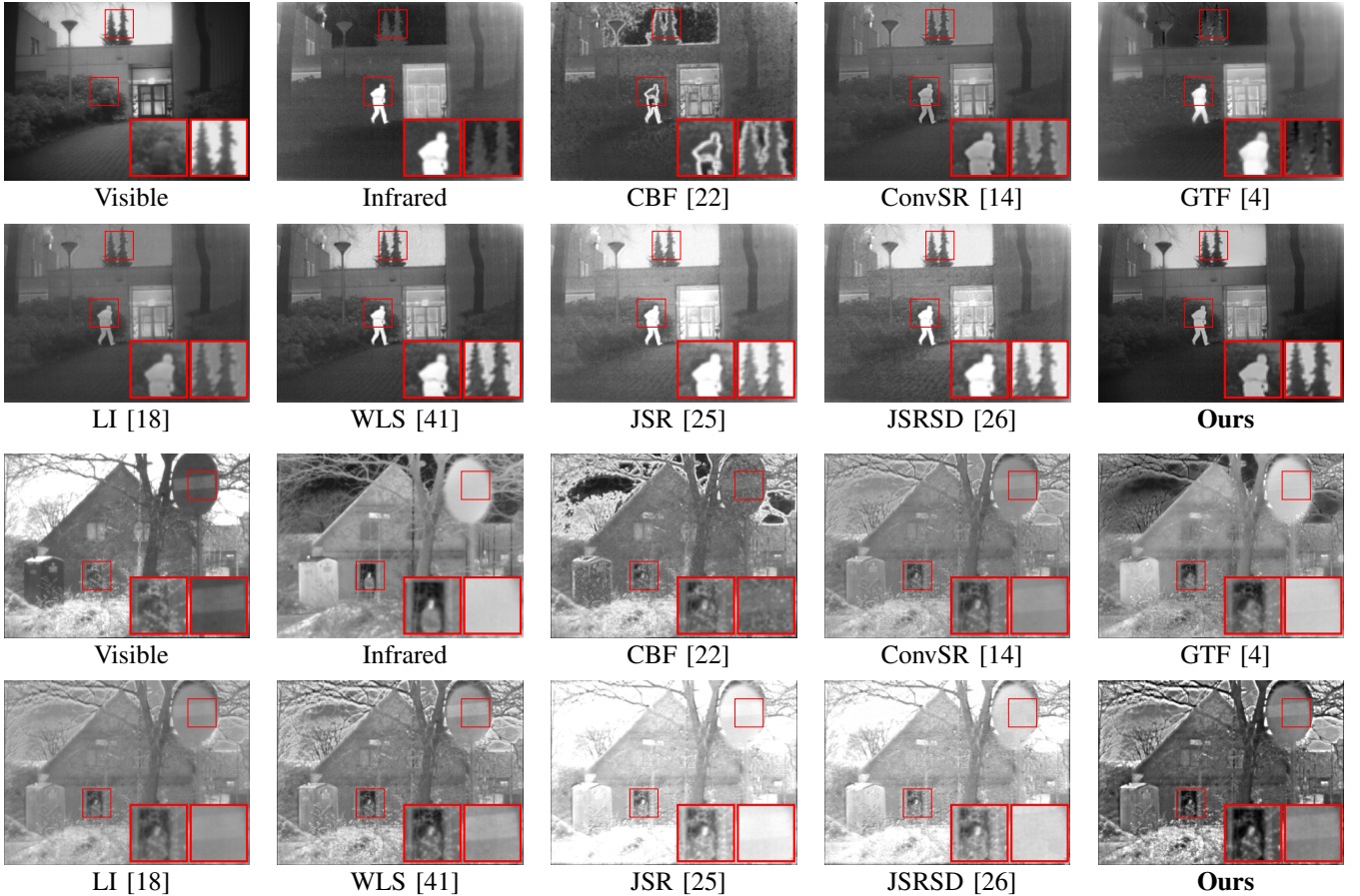
\begin{figure*}[ht!]
\centering
\newcommand{\dwi}{0.18\linewidth}
\newcommand{\negskip}{\noalign{\vspace{-0.2cm}}}
\newcommand{\fig}[5]{
	\begin{tikzpicture}
	\begin{scope}[spy using outlines=
	{magnification=2, size=0.75cm}]
	\node { \includegraphics[width=\dwi]{#1} };
	\spy [red] on #2 in node [left] at #3;
	\spy [red] on #4 in node [left] at #5;
	\end{scope}
	\end{tikzpicture}
}
\newcommand{\figone}[1]{\fig{#1}{(-0.1,0.9)}{(1.6,-0.75)}{(-0.3,0)}{(0.8,-0.75)}}

\newcommand{\figtwo}[1]{\fig{#1}{(1,0.8)}{(1.6,-0.8)}{(-0.4,-0.4)}{(0.8,-0.8)}}

\begin{tabular}{@{} *{5}{@{\hskip 0.001\linewidth} c}}
	\figone{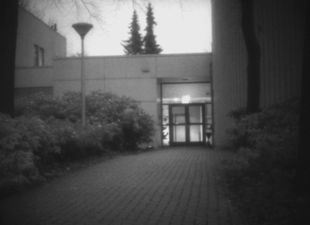} & \figone{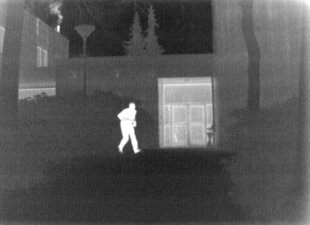} & \figone{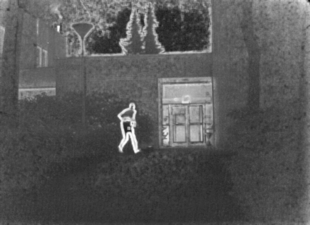} & \figone{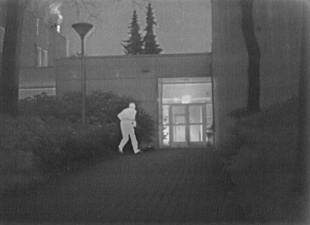} & \figone{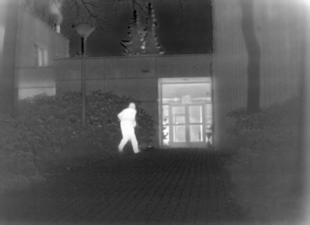} \\
	\negskip
	Visible & Infrared & CBF~\cite{kumar2015image} & ConvSR~\cite{liu2016image} & GTF~\cite{ma2016infrared}\\
	
	\figone{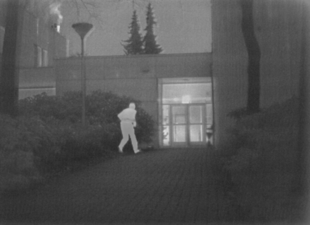} &
	\figone{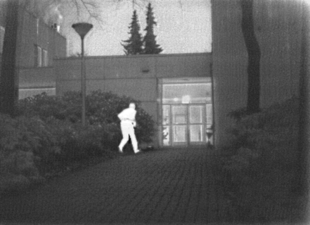} & \figone{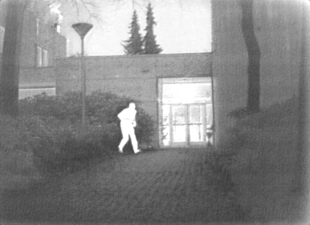} & \figone{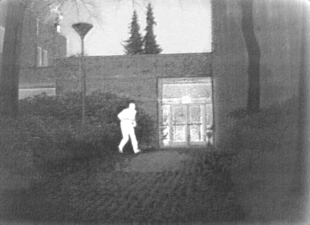} & \figone{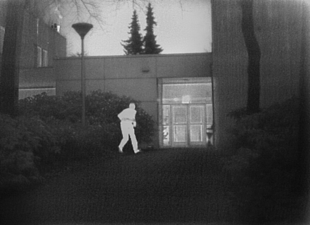} \\
	\negskip
	LI~\cite{li2018infrared} & WLS~\cite{ma2017infrared} & JSR~\cite{zhang2013dictionary} & JSRSD~\cite{liu2017infrared} & \textbf{Ours}\\
	\noalign{\smallskip}
	
	\figtwo{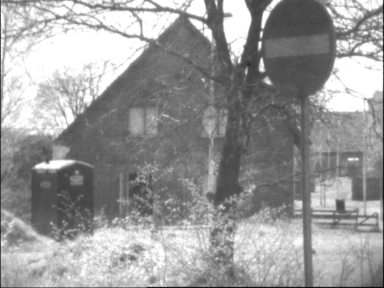} & \figtwo{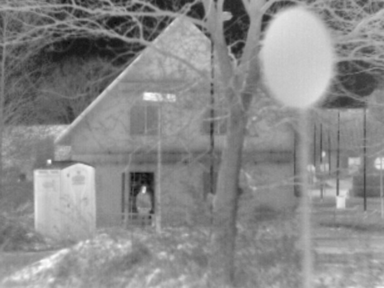} & \figtwo{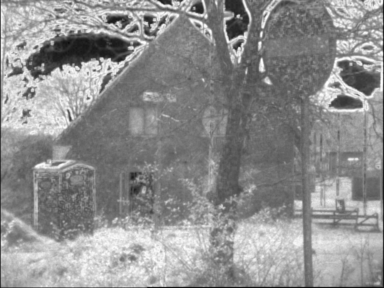} & \figtwo{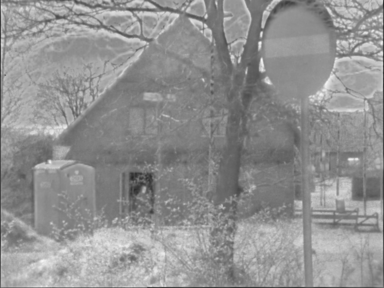} & \figtwo{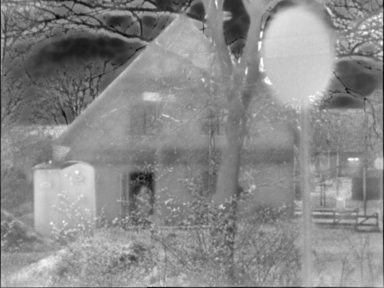} \\
	\negskip
	Visible & Infrared & CBF~\cite{kumar2015image} & ConvSR~\cite{liu2016image} & GTF~\cite{ma2016infrared}\\
	
	\figtwo{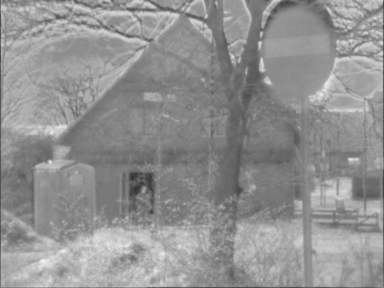} &
	\figtwo{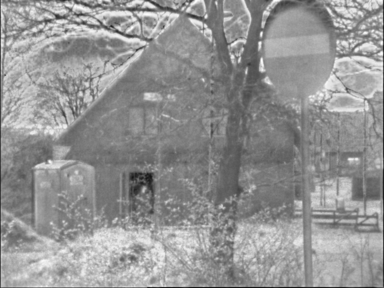} & \figtwo{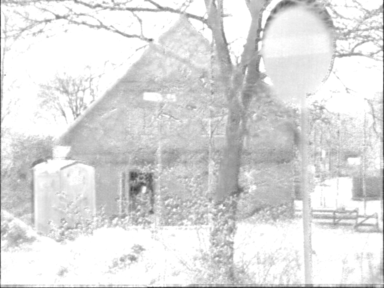} & \figtwo{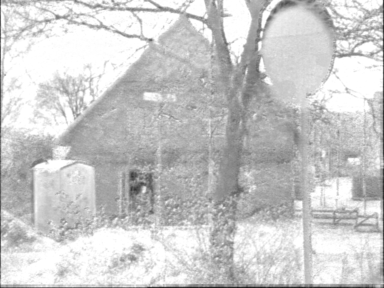} & \figtwo{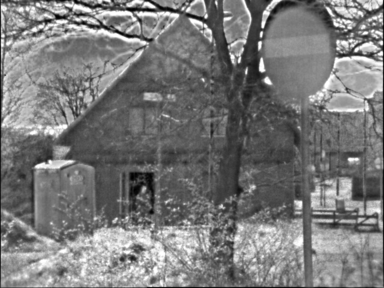} \\
	\negskip
	LI~\cite{li2018infrared} & WLS~\cite{ma2017infrared} & JSR~\cite{zhang2013dictionary} & JSRSD~\cite{liu2017infrared} & \textbf{Ours}\\
	
\end{tabular}

\caption{\label{fig:comparison-infrared} Visible and infrared source images with the fusion results obtained by different methods. Insets are magnified $\times 2$. Best viewed on screen.}
\end{figure*}

\newcolumntype{Z}{>{\centering\arraybackslash}m{1.4cm}}
\begin{table*}[ht]
	\begin{center}
		\caption{\label{tbl:infrared} Objective assessment of different methods on infrared and visible image fusion. \textbf{Bold} and \underline{underlined} values indicate the \textbf{best} and \underline{second-best} scores, respectively. Time was computed on images with an average size of $460\times610$ ($\pm 115\times155$).}
		\begin{tabular}{l *{8}{Z}}
			\hline\noalign{\smallskip}
			Metric & CBF~\cite{kumar2015image} & ConvSR~\cite{liu2016image} & GTF~\cite{ma2016infrared} & LI~\cite{li2018infrared} & WLS~\cite{ma2017infrared} & JSR~\cite{zhang2013dictionary} & JSRSD~\cite{liu2017infrared} & Ours \\ 
			\hline\noalign{\smallskip}
			
			\textit{EN} & \underline{6.857} & 6.259 & 6.635 & 6.182 & 6.638 & 6.363 & 6.693 & \textbf{7.126} \\
			\textit{MI} & \underline{13.714} & 12.517 & 13.271 & 12.364 & 13.276 & 12.727 & 13.386 & \textbf{14.252} \\
			\textit{VIFF} & 0.265 & 0.272 & 0.188 & 0.259 & \underline{0.444} & 0.363 & 0.292 & \textbf{0.690} \\
			\textit{$Q_{MI}$} & \underline{2.039} & 1.946 & 2.006 & 1.934 & 2.006 & 1.963 & 2.014 & \textbf{2.064} \\
			\textit{$Q_G$} & 0.378 & 0.491 & 0.421 & 0.364 & \textbf{0.509} & 0.308 & 0.265 & \underline{0.500} \\
			\textit{$Q_Y$} & 0.643 & 0.802 & 0.726 & 0.702 & \underline{0.805} & 0.588 & 0.501 & \textbf{0.816} \\
			\textit{$Q_C$} & 0.486 & 0.594 & 0.468 & \underline{0.606} & 0.601 & 0.467 & 0.427 & \textbf{0.649} \\
			\textit{$Q_P$} & 0.147 & \textbf{0.355} & 0.205 & 0.297 & 0.309 & 0.173 & 0.118 & \underline{0.315} \\
			
			\hline \noalign{\smallskip}
			Time & 15.28 & 86.8 & 2.57 & 6.20 & \underline{1.28} & 346.18 & 396.74 & \textbf{0.16} \\
			Std $\sigma$ & 6.69 & 37.44 & 1.31 & 2.79 & \underline{0.65} & 151.25 & 178.73 & \textbf{0.08} \\	
			\hline
		\end{tabular}
	\end{center}
\end{table*}

\subsubsection{Architecture and depth impact}
In this experiment, we study the effect of the neural network architecture on the fusion quality. We consider seven models pre-trained on ImageNet~\cite{deng2009imagenet} VGG-16~\cite{simonyan2014very}, AlexNet~\cite{krizhevsky2012imagenet}, ResNet50~\cite{he2016deep}, DenseNet121\cite{huang2017densely}, MobileNet~\cite{howard2017mobilenets}, MobileNetV2~\cite{sandler2018mobilenetv2}, and SqueezeNet~\cite{iandola2016squeezenet}. VGG-16 and AlexNet are feed-forward convolutional neural networks, while ResNet50 and DenseNet121 contain residual connections, i.e.\ direct links from earlier to later stages of the model. Finally, MobileNet, MobileNetV2, and SqueezeNet are smaller and more efficient models designed for limited hardware devices and embedded systems. Finally, we also consider a non trained ResNet50 network, i.e., its parameters are randomly initialized.

Deep neural networks consist of multiple layers, and it is important for our application to understand the effect of depth on the output fusion. To that end, we compute the image fusion at multiple depth levels of the network and compare them. Here, we define network depth as the layer at the end of a convolution block. For example, the VGG-16 architecture consists of $5$ convolution blocks, and we take the features at the end of each one. We proceed similarly for the other architectures. For more information on their structure, please refer to the original works~\cite{simonyan2014very,krizhevsky2012imagenet,he2016deep,huang2017densely,howard2017mobilenets,sandler2018mobilenetv2,iandola2016squeezenet}.

%auto-ignore
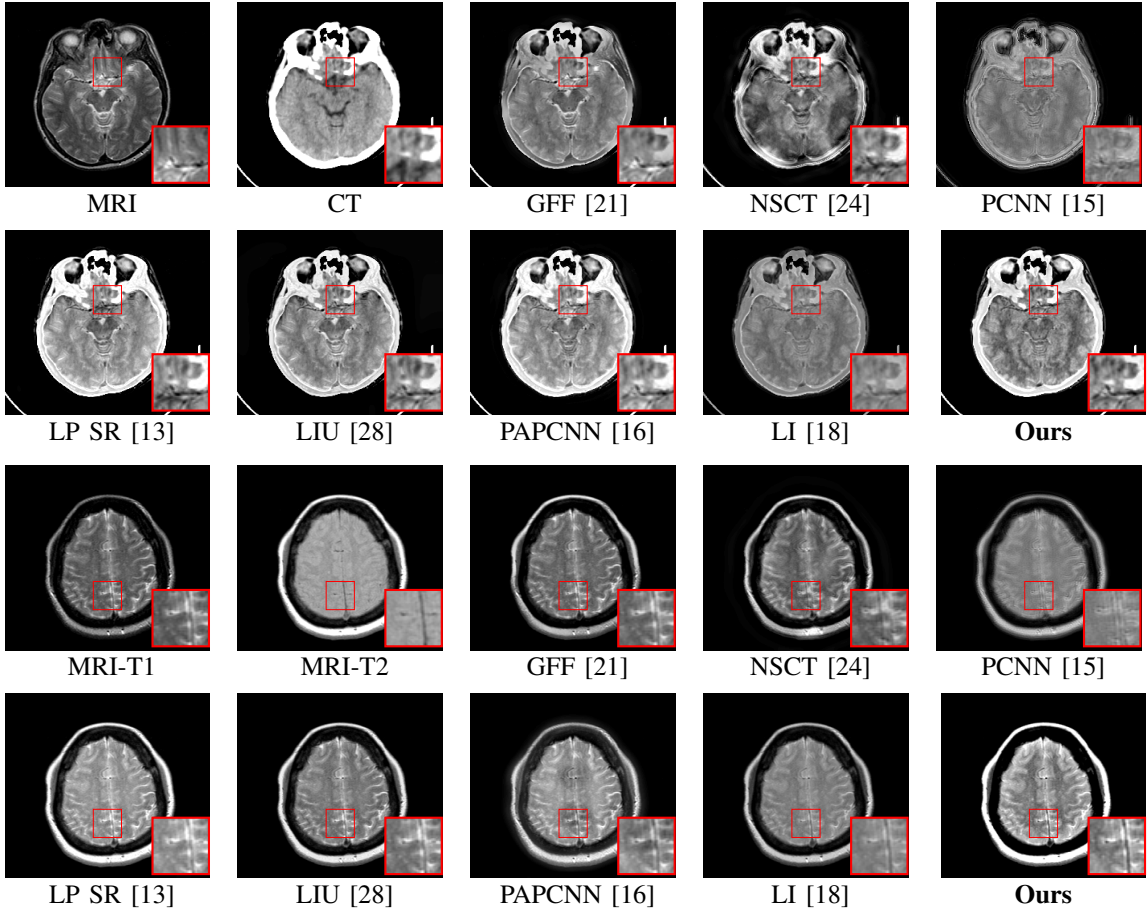
\begin{figure*}[ht!]
\centering
\newcommand{\dwi}{0.15\linewidth}
\newcommand{\dhi}{0.135\linewidth}
\newcommand{\negskip}{\noalign{\vspace{-0.2cm}}}
\newcommand{\fig}[3]{
	\begin{tikzpicture}
	\begin{scope}[spy using outlines=
	{magnification=2, size=.75cm}]
	\node { \includegraphics[width=\dwi,height=\dhi]{#1} };
	\spy [red] on #2 in node [left] at #3;
	\end{scope}
	\end{tikzpicture}
}
\newcommand{\figone}[1]{\fig{#1}{(0,0.3)}{(1.35,-0.8)}}
\newcommand{\figtwo}[1]{\fig{#1}{(0,-0.5)}{(1.35,-0.8)}}

\begin{tabular}{@{} *{5}{@{\hskip 0.001\linewidth} c}}
	\figone{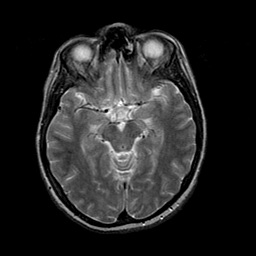} & \figone{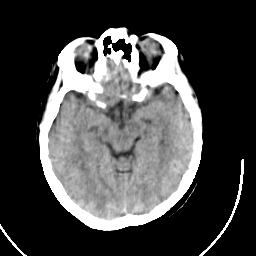} &
	\figone{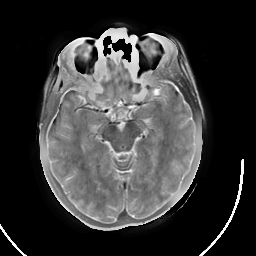} &
	\figone{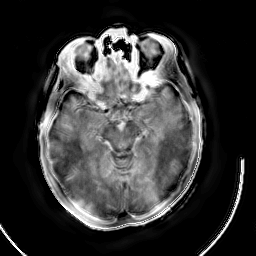} &
	\figone{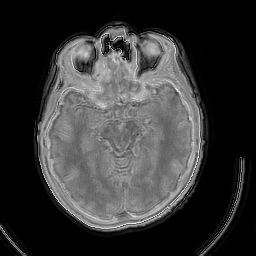} \\
	\negskip
	MRI & CT & GFF~\cite{li2013image} & NSCT~\cite{bhatnagar2013directive} & PCNN~\cite{wang2013multimodal}\\
	
	\figone{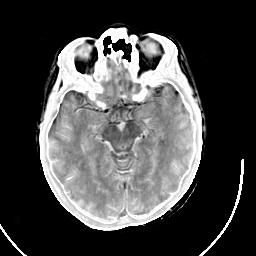} & \figone{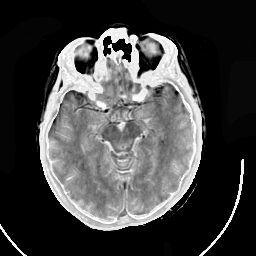} &
	\figone{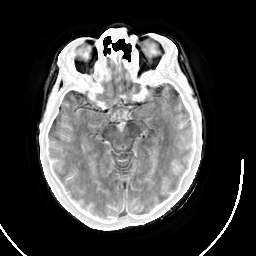} &
	\figone{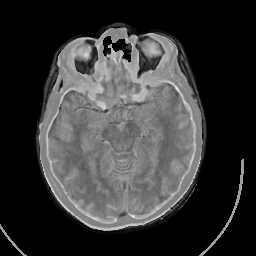} &
	\figone{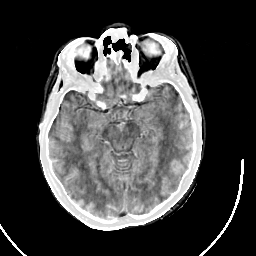}\\
	\negskip
	LP SR~\cite{liu2015general} & LIU~\cite{liu2017medical} & PAPCNN~\cite{yin2018medical} & LI~\cite{li2018infrared} & \textbf{Ours}\\
	\noalign{\smallskip}
	
	\figtwo{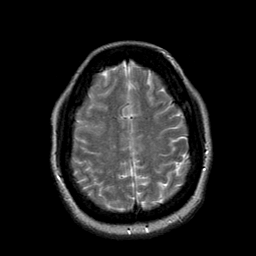} & \figtwo{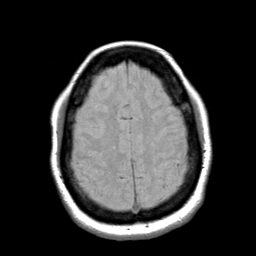} &
	\figtwo{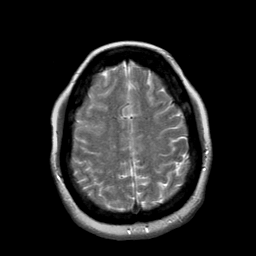} &
	\figtwo{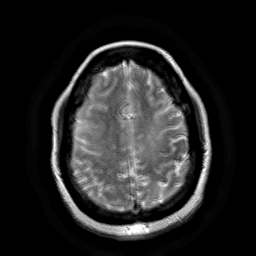} &
	\figtwo{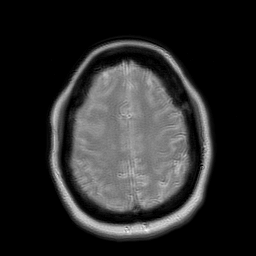} \\
	\negskip
	MRI-T1 & MRI-T2 & GFF~\cite{li2013image} & NSCT~\cite{bhatnagar2013directive} & PCNN~\cite{wang2013multimodal}\\
	
	\figtwo{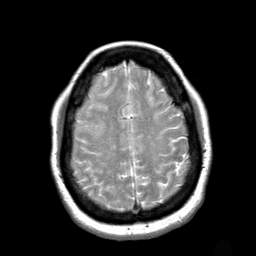} & \figtwo{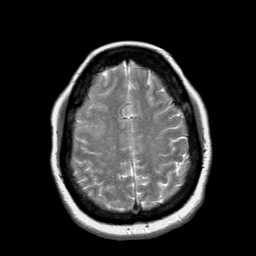} &
	\figtwo{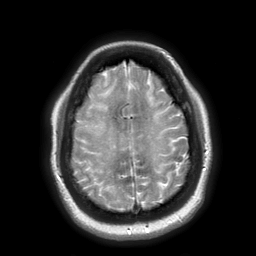} &
	\figtwo{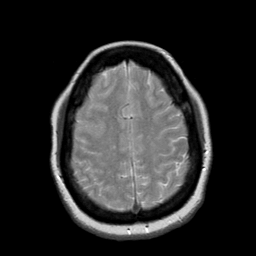} &
	\figtwo{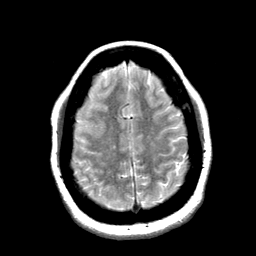}\\
	\negskip
	LP SR~\cite{liu2015general} & LIU~\cite{liu2017medical} & PAPCNN~\cite{yin2018medical} & LI~\cite{li2018infrared} & \textbf{Ours}\\
	\noalign{\smallskip}
	
\end{tabular}

\caption{\label{fig:comparison-medical} MRI-T1 and MRI-T2 source images with the fusion results obtained by different methods. Insets are magnified $\times 2$. Best viewed on screen.}
\end{figure*}

\begin{table*}[ht]
	\begin{center}
		\caption{\label{tbl:medical} Objective assessment of different methods on medical image fusion (MRI-CT and T1-T2). \textbf{Bold} and \underline{underlined} values indicate the \textbf{best} and \underline{second-best} scores, respectively. Time was computed on images with a size of $256\times256$.}
		\begin{tabular}{l *{8}{Z}}
			\hline \noalign{\smallskip}
			
			Metric & GFF~\cite{li2013image} & NSCT~\cite{bhatnagar2013directive} & PCNN~\cite{wang2013multimodal} & LP SR~\cite{liu2015general} & LIU~\cite{liu2017medical} & PAPCNN~\cite{yin2018medical} & LI~\cite{li2018infrared} & Ours \\ 
			\hline\noalign{\smallskip}
			\textit{EN} & 3.322 & 3.579 & 3.481 & 3.424 & 3.610 & \underline{3.969} & 3.329 & \textbf{4.470} \\
			\textit{MI} & 6.644 & 7.158 & 6.963 & 6.849 & 7.219 & \underline{7.937} & 6.657 & \textbf{8.939} \\
			\textit{VIFF} & 0.110 & 0.110 & 0.085 & \underline{0.164} & 0.144 & 0.112 & 0.108 & \textbf{0.752} \\
			\textit{$Q_{MI}$} & 0.953 & 0.694 & 0.787 & \underline{0.976} & 0.899 & 0.761 & \textbf{1.022} & 0.882 \\
			\textit{$Q_G$} & 0.488 & 0.399 & 0.447 & \underline{0.496} & 0.437 & 0.435 & 0.481 & \textbf{0.710} \\
			\textit{$Q_Y$} & 0.539 & 0.405 & 0.447 & \underline{0.565} & 0.531 & 0.513 & 0.534 & \textbf{0.826} \\
			\textit{$Q_C$} & 0.508 & 0.370 & 0.412 & \underline{0.522} & 0.494 & 0.477 & 0.491 & \textbf{0.744} \\
			\textit{$Q_P$} & \textbf{0.720} & 0.312 & 0.105 & 0.624 & 0.638 & 0.524 & 0.589 & \underline{0.639} \\
			
			\hline \noalign{\smallskip}
			Time & 0.05 & 3.49 & 0.46 & \underline{0.04} & 13.12 & 6.92 & 1.65 & \textbf{0.03} \\
			Std $\sigma$ & 0.02 & 1.90 & 0.10 & \underline{0.008} & 1.91 & 2.46 & 0.22 & \textbf{0.001} \\
			\hline
		\end{tabular}
	\end{center}
\end{table*}

Fig.~\ref{fig:net-depth} shows the average fusion metrics on images generated using the compared network architectures at different depths. All pre-trained networks perform similarly on the evaluated metrics. The standard deviations of the mean metric across different pre-trained networks are $2\times10^{-3}$ for $Q_G$, $1\times10^{-3}$ for $Q_Y$, $5\times10^{-3}$ for $Q_C$, and $3\times10^{-3}$ for $Q_P$, showing the similarity between those results. These experiments highlight that our method is stable under network and depth variations, and can be applied using a variety of networks. Additionally, comparing the quality of pre-trained and non trained ResNet50 on all depth levels shows the importance of using pre-trained models that have learned natural image representations. Having similar performance than the feed-forward architectures, the residual networks offer lower memory constraints. Additionally, MobileNet, MobileNetV2, and SqueezeNet perform as well as the other architectures, and can be adopted on low-energy and limited hardware systems.

%auto-ignore
\subsection{Comparison with other methods}
We compare our method with different fusion algorithms on thermal, medical, and multi-focus fusion. We select commonly used fusion methods pertaining to each fusion task. The parameters of all the evaluated methods are set to the default values from their publicly available code. Following the observations from the ablation studies, we set the following configuration for our method. We use a box filter for the two-scale decomposition and set the guided filter parameters to $r_b=45$, $\epsilon_b=0.1$, $r_d=7$, and $\epsilon_d=10^{-6}$. We generate the detail weight maps using a ResNet50 network, with feature maps extracted at depth $l=3$.

%auto-ignore
\begin{figure*}[ht!]
\centering
\newcommand{\dwi}{0.15\linewidth}
\newcommand{\dhi}{0.12\linewidth}
\newcommand{\negskip}{\noalign{\vspace{-0.2cm}}}
\newcommand{\fig}[3]{
	\begin{tikzpicture}
	\begin{scope}[spy using outlines=
	{magnification=4, size=1cm}]
	\node { \includegraphics[width=\dwi,height=\dhi]{#1} };
	\spy [red] on #2 in node [left] at #3;
	\end{scope}
	\end{tikzpicture}
}
\newcommand{\figone}[1]{\fig{#1}{(-0.2,-0.5)}{(1.35,-0.58)}}
\newcommand{\figtwo}[1]{\fig{#1}{(0.2,0.4)}{(1.35,-0.58)}}
\newcommand{\figthree}[1]{\fig{#1}{(0.6,0.85)}{(1.35,-0.58)}}
\newcommand{\fnum}{13}
\newcommand{\focusn}{7}
\begin{tabular}{@{} *{5}{@{\hskip 0.001\linewidth} c}}
	\figtwo{images/focus/f1/\fnum.png} & \figtwo{images/focus/f2/\fnum.png} &
	\figtwo{images/focus/ConvSR/\fnum.png} &
	\figtwo{images/focus/CBF/\fnum.png} &
	\figtwo{images/focus/GFF/\fnum.png} \\
	\negskip
	Focus 1 & Focus 2 & ConvSR~\cite{liu2016image} & CBF~\cite{kumar2015image} & GFF~\cite{li2013image}\\
	
	\figtwo{images/focus/NSCT/\fnum.png} & \figtwo{images/focus/CNN/\fnum.png} &
	\figtwo{images/focus/DSIFT/\fnum.png} &
	\figtwo{images/focus/BF/\fnum.png} &
	\figtwo{images/focus/Ours/\fnum.png} \\
	\negskip
	NSCT~\cite{bhatnagar2013directive} & LIU-M~\cite{liu2017multi} & DSIFT~\cite{liu2015multi} & BFF~\cite{zhang2017boundary} & \textbf{Ours}\\
	\noalign{\smallskip}
	
	\figone{images/focus/f1/\focusn.png} & \figone{images/focus/f2/\focusn.png} &
	\figone{images/focus/ConvSR/\focusn.png} &
	\figone{images/focus/CBF/\focusn.png} &
	\figone{images/focus/GFF/\focusn.png} \\
	\negskip
	Focus 1 & Focus 2 & ConvSR~\cite{liu2016image} & CBF~\cite{kumar2015image} & GFF~\cite{li2013image}\\
	
	\figone{images/focus/NSCT/\focusn.png} & \figone{images/focus/CNN/\focusn.png} &
	\figone{images/focus/DSIFT/\focusn.png} &
	\figone{images/focus/BF/\focusn.png} &
	\figone{images/focus/Ours/\focusn.png} \\
	\negskip
	NSCT~\cite{bhatnagar2013directive} & LIU-M~\cite{liu2017multi} & DSIFT~\cite{liu2015multi} & BFF~\cite{zhang2017boundary} & \textbf{Ours}\\
	\noalign{\smallskip}
	
%	\figthree{images/lytro/f1/lytro-0\focusn-A.png} & \figthree{images/lytro/f2/lytro-0\focusn-B.png} &
%	\figthree{images/lytro/ConvSR/\focusn.png} &
%	\figthree{images/lytro/CBF/\focusn.png} &
%	\figthree{images/lytro/GFF/\focusn.png} \\
%	\negskip
%	Focus 1 & Focus 2 & ConvSR~\cite{liu2016image} & CBF~\cite{kumar2015image} & GFF~\cite{li2013image}\\
%	
%	\figthree{images/lytro/NSCT/\focusn.png} & \figthree{images/lytro/CNN/\focusn.png} &
%	\figthree{images/lytro/DSIFT/\focusn.png} &
%	\figthree{images/lytro/BF/\focusn.png} &
%	\figthree{images/lytro/Ours/\focusn.png} \\
%	\negskip
%	NSCT~\cite{bhatnagar2013directive} & LIU-M~\cite{liu2017multi} & DSIFT~\cite{liu2015multi} & BF~\cite{zhang2017boundary} & \textbf{Ours}\\
%	\noalign{\smallskip}
	
\end{tabular}

\caption{\label{fig:comparison-focus} Multi-focus source images with the fusion results obtained by different methods. Insets are magnified $\times 4$. Best viewed on screen.}
\end{figure*}
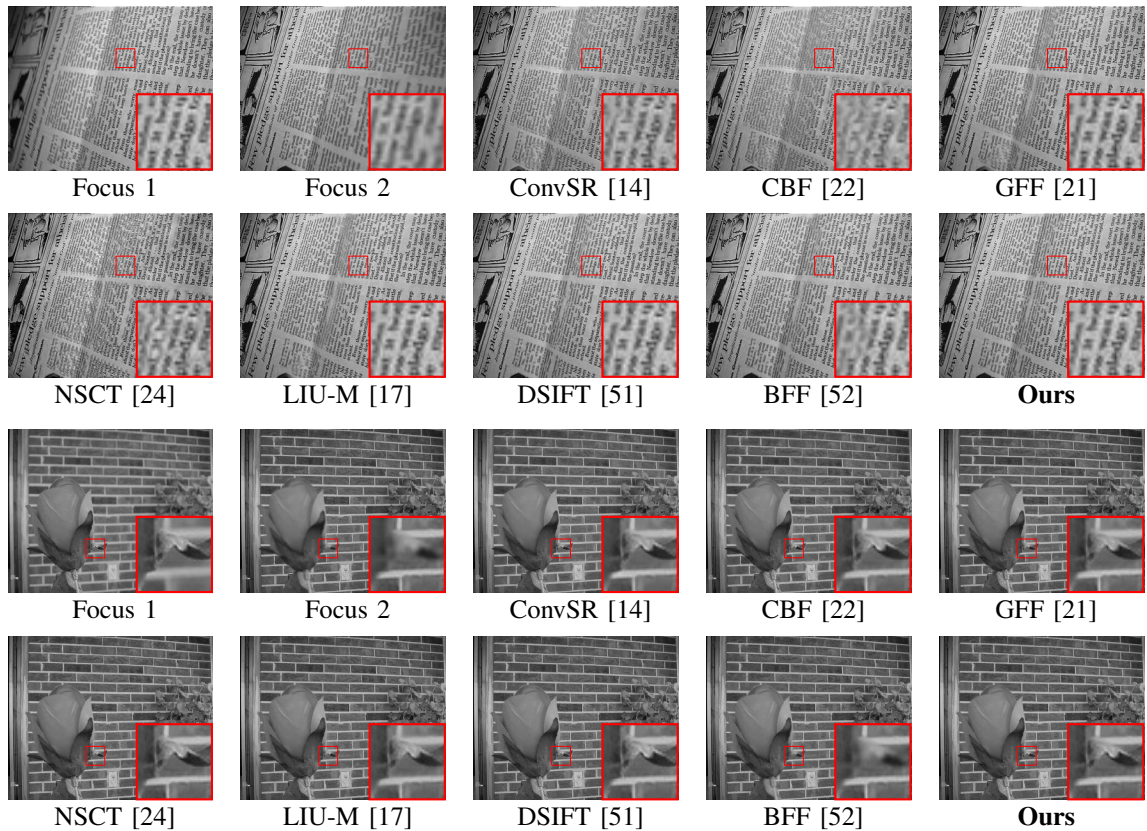

\begin{table*}[ht]
	\begin{center}
		\caption{\label{tbl:focus} Objective assessment of different methods on multi-focus image fusion. \textbf{Bold} and \underline{underlined} values indicate the \textbf{best} and \underline{second-best} scores, respectively. Time was computed on images with an average size of $360\times450$ ($\pm 135\times155$).}
		\begin{tabular}{l *{8}{Z}}
			\hline \noalign{\smallskip}
			Metric &ConvSR~\cite{liu2016image} & CBF~\cite{kumar2015image} & GFF~\cite{li2013image} & NSCT~\cite{bhatnagar2013directive} & LIU-M~\cite{liu2017multi} & DSIFT~\cite{liu2015multi} & BFF~\cite{zhang2017boundary} & Ours \\ 
			\hline\noalign{\smallskip}
			\textit{EN} & 7.261 & 7.281 & 7.286 & \textbf{7.297} & 7.279 & 7.280 & 7.275 & \underline{7.287} \\
			\textit{MI} & 14.522 & 14.562 & 14.573 & \textbf{14.595} & 14.559 & 14.560 & 14.551 & \underline{14.575} \\
			\textit{VIFF} & 0.840 & 0.876 & 0.886 & \underline{0.896} & 0.885 & 0.887 & 0.881 & \textbf{0.899} \\
			\textit{$Q_{MI}$} & 0.869 & 0.996 & 1.048 & 0.931 & 1.118 & \underline{1.148} & 1.143 & \textbf{1.155} \\
			\textit{$Q_{G}$} & 0.588 & 0.677 & 0.703 & 0.673 & \textbf{0.710} & \underline{0.709} & \textbf{0.710} & 0.708 \\
			\textit{$Q_y$} & 0.888 & 0.950 & 0.975 & 0.955 & \underline{0.985} & 0.982 & \textbf{0.987} & \underline{0.985} \\
			\textit{$Q_c$} & 0.639 & \textbf{0.659} & 0.642 & \underline{0.650} & 0.635 & 0.629 & 0.629 & 0.646 \\
			\textit{$Q_p$} & 0.794 & 0.810 & 0.834 & 0.806 & \underline{0.839} & 0.836 & 0.836 & \textbf{0.841} \\
			
			\hline \noalign{\smallskip}
			Time & 62.70 & 10.33 & \underline{0.14} & 45.87 & 80.79 & 4.24 & 0.58 & \textbf{0.08} \\
			Std $\sigma$ & 40.07 & 6.58 & \underline{0.09} & 26.31 & 38.08 & 4.01 & 0.77 & \textbf{0.05} \\
			\hline
		\end{tabular}
	\end{center}
\end{table*}

\subsubsection{Infrared and visible fusion}
The methods we compare against are CBF~\cite{kumar2015image}, ConvSR~\cite{liu2016image}, GTF~\cite{ma2016infrared}, WLS~\cite{ma2017infrared}, JSR~\cite{zhang2013dictionary}, JSRSD~\cite{liu2017infrared}, and LI~\cite{li2018infrared}. CBF uses a cross bilateral filter to extract detail layers of source images, and use those layers to generate weight maps to combine the sources. ConvSR uses convolutional sparse representation to fuse the detail layers of two-scale decomposed images, while averaging the base layers.  GTF poses the infrared and visible image fusion as total variation minimization problem in the infrared domain constrained to the gradients of the visible image. WLS uses visual saliency and weighted least-squares to fuse two-scale image decompositions. JSR and JSRSD use dictionaries to learn sparse image representations for weight computation, JSRSD additionally uses saliency information to guide the fusion. LI uses a neural network to directly fuse the high frequencies and averages the low frequencies.

Fig.~\ref{fig:comparison-infrared} shows two pairs of visible and infrared images, and the fusion results from the compared methods. In comparison with CBF and ConvSR, our method has less edge and ringing artifacts. Additionally, JSR and JSRSD produce over-exposed images while LI generates under-exposed results. In GTF, it's very difficult to notice the trees, as seen in the inset image. In contrast, our method better respects the intensity values present in the source images. Additionally, in the second pair of images, the person and the sign post are most noticeable in our fusion among all the compared methods. 

Additionally, Table~\ref{tbl:infrared} summarizes the average objective metrics on infrared and visible fusion of all images in the TNO dataset. Our method consistently records the highest or second-highest performance on all the evaluated metrics, achieving state-of-the-art performance on thermal fusion.

\subsubsection{Medical image fusion}
For this task, the compared methods are GFF~\cite{li2013image}, NSCT~\cite{bhatnagar2013directive}, PCNN~\cite{wang2013multimodal}, LP SR~\cite{liu2015general}, LIU~\cite{liu2017medical}, PAPCNN~\cite{yin2018medical}, and LI~\cite{li2018infrared}. GFF is based on a base and detail decomposition with guided filters to generate weights. NSCT and PCNN both use the non-subsampled contourlet transform to decompose the images into low and high frequencies, and apply different fusion rules depending on the frequency. LP SR uses Laplacian pyramids to decompose the images and applies sparse representation for low frequency fusion and maximum selection rule for high frequency fusion.
LIU employs Siamese Neural Networks~\cite{bromley1994signature} to predict the fusion weights with input images in the spatial domain. PAPCNN uses the non-subsampled shearlet transform for image decomposition, with an energy based fusion for low frequencies, and an adaptive PCNN for high frequencies. LI is the same method presented in the previous section.

Fig.~\ref{fig:comparison-medical} shows two pairs of images corresponding to grayscale fusion of MRI-CT and T1-T2 images, respectively. GFF, NSCT and PAPCNN are not able to capture complementary information in all the image regions, which is reflected in the stark contrast inside regions that should have similar intensities. This can be seen for GFF in the MRI-CT fusion, where some edges exhibit a single intensity in both sources but present different levels in the fusion. Similar artifacts can be spotted for NSCT in MRI-CT and PAPCNN in T1-T2 fusion. In contrast, our fusion approach relies on neural networks that have been trained on various intensity levels and thus respects the intensity differences inside and between salient regions leading to a more intensity consistent fusion that is easier to interpret. Additionally, LIU and LI generate weights that are highly biased towards the highest intensity signal. This results in low contrast images, where darker details from sources do not appear in the fusion. In comparison, our method better represents edges between neighboring regions while respecting their saliency, resulting in higher contrast, noise-free fusions. Finally, Table~\ref{tbl:medical} shows the performance of different methods averaged over both the MRI-CT and T1-T2 fusion sets. The evaluations show that our method pushes state-of-the-art performance on all metrics except in $Q_{MI}$ where it still obtains competitive results.

\begin{figure*}[ht!]
\centering
\newcommand{\dwi}{0.27\textwidth}
\newcommand{\dhi}{0.18\textwidth}
\newcommand{\negskip}{\noalign{\vspace{0cm}}}
\newcommand{\fig}[3]{
	\includegraphics[width=\dwi,height=\dhi]{#1}
}
\newcommand{\figone}[1]{\fig{#1}{(0,0)}{(1.35,-0.8)}}
\newcommand{\figtwo}[1]{\fig{#1}{(0,0.1)}{(1.35,-0.8)}}

\newcolumntype{T}{ >{\centering\arraybackslash} m{0.6cm} }
\newcolumntype{P}{ >{\centering\arraybackslash} m{0.3\textwidth} }
\begin{minipage}{0.48\textwidth}
\begin{tabular}{T *{3}{@{\hskip 0.01em} P}}
	
	$I_k$ & \figone{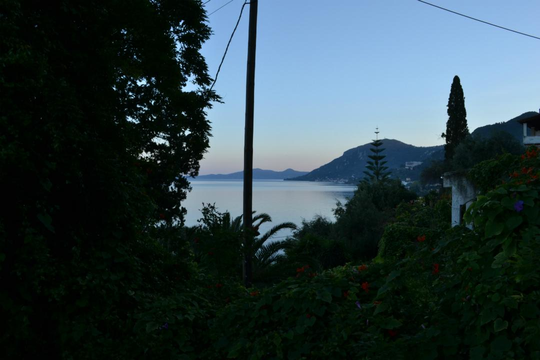} & \figone{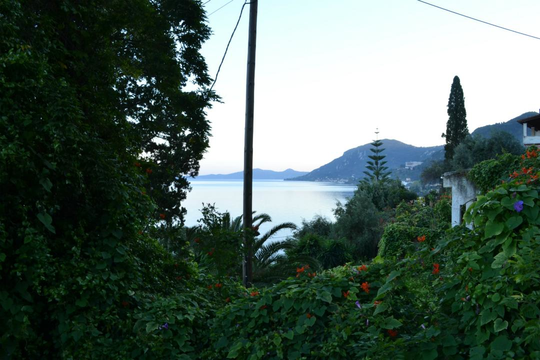} &
	\figone{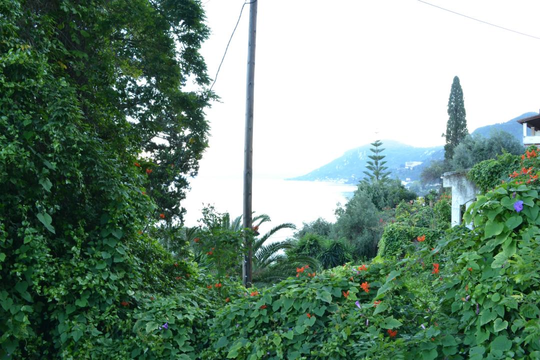} \\
	\negskip
	
	$W^B_k$ & \figone{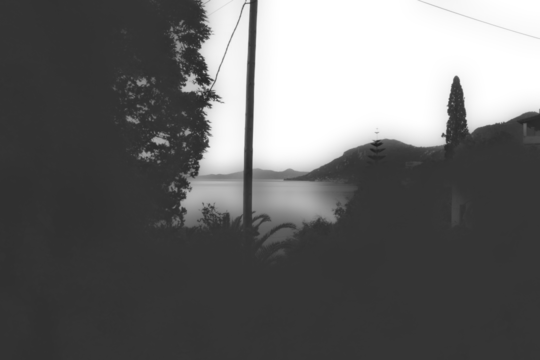} & \figone{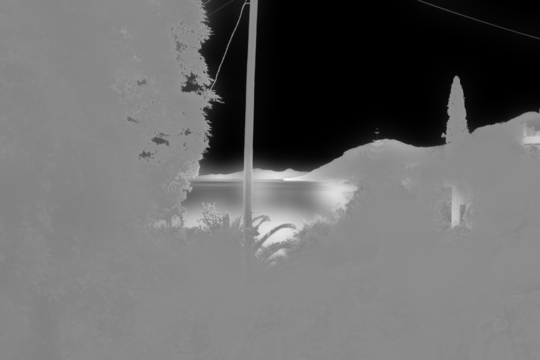} &
	\figone{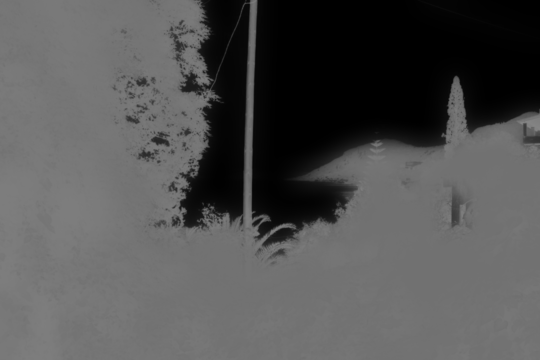} \\
	\negskip
	
	$W^D_k$ & \figone{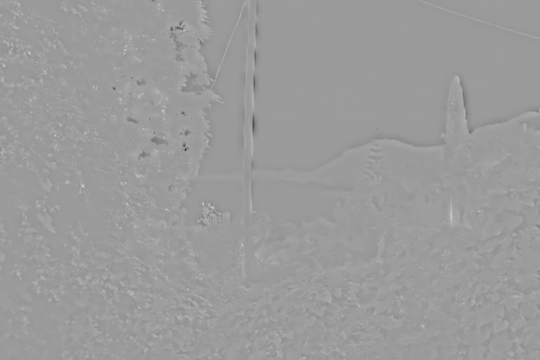} & \figone{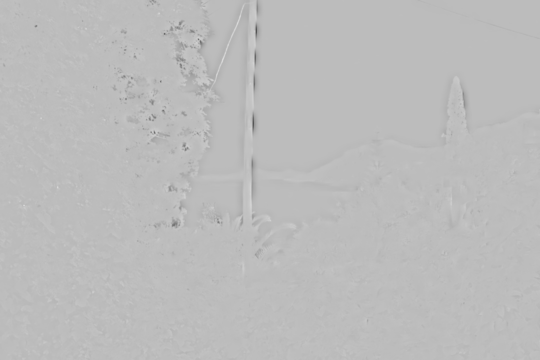} &
	\figone{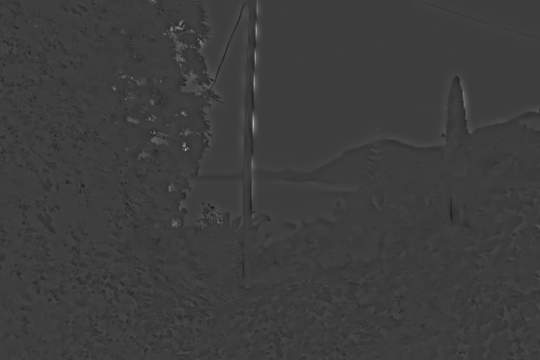} \\
	\noalign{\medskip}
	
	$I_k$ & \figone{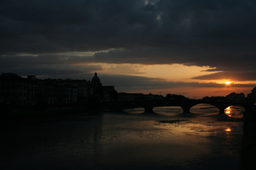} & \figone{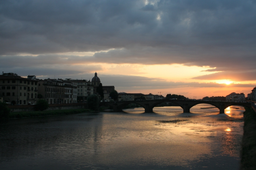} &
	\figone{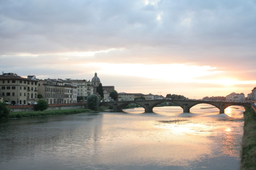} \\
	\negskip
	
	$W^B_k$ & \figone{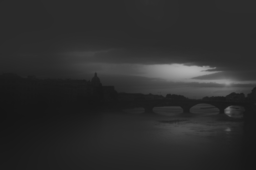} & \figone{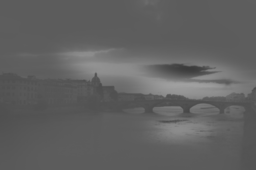} &
	\figone{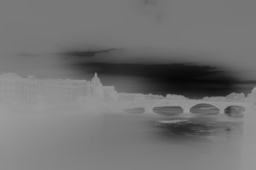} \\
	\negskip
	
	$W^D_k$ & \figone{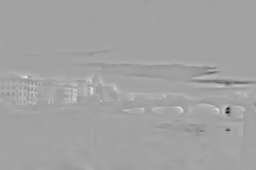} & \figone{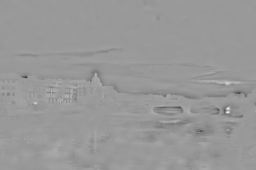} &
	\figone{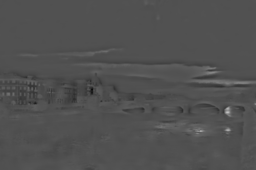} \\
	\noalign{\smallskip}
	
\end{tabular}
\end{minipage}
\begin{minipage}{0.49\textwidth}
	\begin{tabular}{c}
		\includegraphics[width=\textwidth, height=0.558\textwidth]{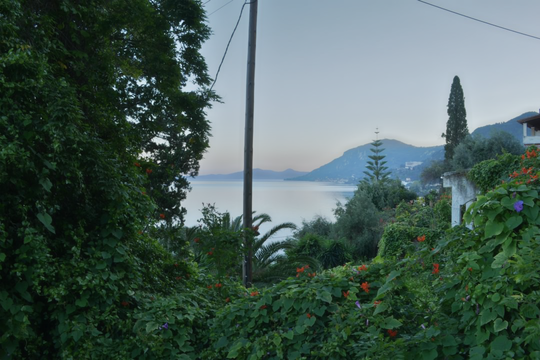} \\
		
		\noalign{\medskip}
		
		\includegraphics[width=\textwidth, height=0.558\textwidth]{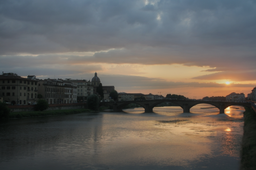} \\
		
		\noalign{\smallskip}
	\end{tabular}
\end{minipage}

\caption{\label{fig:multi-multi} Multi-input fusion examples. On the left side, the images are organized as sources (rows 1 and 4), base weights (rows 2 and 5), and detail weights (rows 3 and 6). On the right side, the resulting image fusion is shown for both sequences. Best viewed on screen.}
\end{figure*}

\subsubsection{Multi-focus fusion}
We compare against the previously mentioned ConvSR~\cite{liu2016image}, CBF~\cite{kumar2015image}, GFF~\cite{li2013image}, and NSCT~\cite{bhatnagar2013directive}. We also compare against LIU-M~\cite{liu2017multi}, DSIFT~\cite{liu2015multi}, and BF (refered to as boundary finding fusion BFF, to reduce confusion with the bilateral filter)~\cite{zhang2017boundary}. LIU-M uses convolutional neural networks to compare image patches and propose weight maps. Then, the weight map boundaries are refined using morphological operators and guided filters. DSIFT uses local feature descriptors to generate an initial decision map and refines it via local matching. Finally, BFF finds and classifies the boundaries between focused and defocused image regions, then generates the fused image by combining the focused parts of the source images. The other fusion methods do not use any boundary adherence checks.

Fig.~\ref{fig:comparison-focus} shows two pairs of multi-focus source images, and their corresponding fusions using the evaluated methods. In the first pair of images, notice that ConvSR, CBF, and NSCT generate artifacts that were not present in any source image. Additionally, BFF is not able to properly estimate the boundaries as can be seen in the inset. GFF, LI, DSIFT, and our method better estimate the boundaries between the focused and unfocused images. Similar observations can be made for the second row, with the flower petal unfocused in the BFF result, while more properly focused in ours. Table~\ref{tbl:focus} summarizes the performance of the evaluated methods on the multi-focus and Lytro datasets. Our method is competitive with DSIFT and BFF, which are the state-of-the-art approaches for multi-focus fusion. While BFF is able to obtain better performance on some metrics, it is unable to always correctly estimate the focus boundaries in the qualitative evaluation.

\subsection{Computational costs}
In addition to the quantitative and qualitative experiments, we conduct runtime evaluations for every compared method. The last two rows of Tables~\ref{tbl:infrared},~\ref{tbl:medical}, and~\ref{tbl:focus} show the average runtime of each method with its standard deviation on images taken from the different datasets. The experiments were run on an Intel Core i7-7700HQ CPU (2.8GHz), and a GeForce GTX 1050 GPU (2Gb) for those methods requiring it (\hspace{-0.01cm}\cite{liu2017medical,liu2017multi,li2018infrared}). All our computations are done on CPU, except the forward pass through the convolutional neural network which is run on GPU. However, thanks to the robustness of our method to different network architectures, we use a small network (ResNet50) whose runtime slows down by only $2\%(\pm0.3)$ when moving its forward pass from GPU to CPU, barely impacting the overall runtime of the proposed pipeline.

Across the three evaluated fusion tasks, and the various source image dimensions, our method has the fastest run time while still obtaining state-of-the-art performance. This is because both saliency and guided filter algorithms run in $O(N)$ time complexity, and they can be even further sped up using GPU implementations~\cite{rahman2011parallel,dai2017hardware,wu2018fast}. These qualities make our approach suitable for deployment on limited hardware architectures.

\subsection{Extension to multiple images}
In this section, we demonstrate the applicability of our method to fusion problems with more than $2$ inputs. For instance, multi-exposure fusion typically require multiple images in the exposure stack to capture the whole dynamic range and minimize the exposure bias difference between consecutive images~\cite{ram2017deepfuse}. Fig.~\ref{fig:multi-multi} shows two multi-exposure image sequences taken from~\cite{ma2018multi} and their respective fusion results using our proposed method. The weight maps show how the base layers preserve the intensity levels while respecting regions boundaries. The detail weight maps also show the consistency with the edges present in the sources, allowing for all details to be preserved in the result, as can be seen on the well-defined leaf edges in the first example. 

This example illustrates how the two-scale decomposition affects the fusion, but more importantly that our method works well even on sequences of input images. Note that the feed-forward pass of neural networks allows batch-processing of inputs. This means that the computational speed is barely affected by the number of images in the input sequence.

\begin{figure}[t]
	\centering
	\newcommand{\dwi}{0.3\textwidth}
	\newcommand{\dhi}{0.25\textwidth}
	\newcommand{\negskip}{\noalign{\vspace{-0.2cm}}}
	\newcommand{\fig}[1]{
		\includegraphics[width=\dwi,height=\dhi]{#1}
	}
	\newcommand{\figr}[3]{
		\begin{tikzpicture}
		\begin{scope}[spy using outlines=
		{magnification=4, size=0.75cm}]
		\node { \includegraphics[width=\dwi,height=\dhi]{#1} };
		\spy [red] on #2 in node [left] at #3;
		\end{scope}
		\end{tikzpicture}
	}
	\newcommand{\figone}[1]{\figr{#1}{(-0.2,0.85)}{(1.35,-0.75)}}
	
	\newcolumntype{T}{ >{\centering\arraybackslash} m{0.24\textwidth} }
	\newcolumntype{P}{ >{\centering\arraybackslash} m{0.32\textwidth} }
	\begin{minipage}{0.5\textwidth}
		\begin{tabular}{*{3}{@{\hskip 0.01em} P}}
			\figone{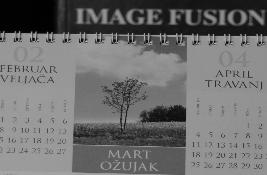} & \figone{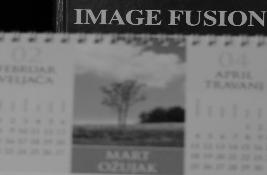} & \figone{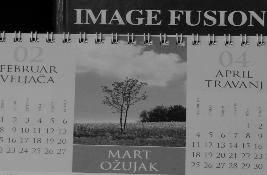}\\
			\negskip
			
			$I_1$ & $I_2$ & $\overline{F}$ \\
			\noalign{\medskip}
		\end{tabular}
	\end{minipage}

	\renewcommand{\dwi}{0.22\textwidth}
	\renewcommand{\dhi}{0.18\textwidth}
	\begin{minipage}{0.5\textwidth}
		\begin{tabular}{*{4}{@{\hskip 0.01em} T}}
			\includegraphics[width=\dwi,height=\dhi,cframe=gray]{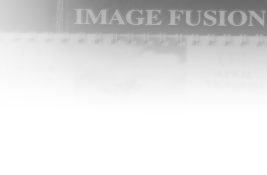}
			& \fig{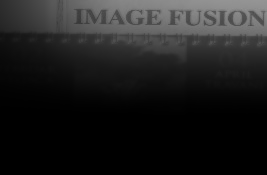} & \fig{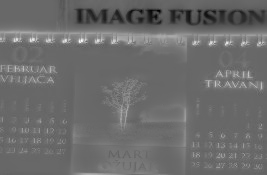} & \fig{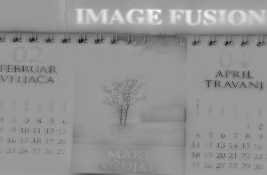}\\
			
			$W^B_1$ & $W^B_2$ & $W^D_1$ & $W^D_2$ \\
			\noalign{\smallskip}
		\end{tabular}
	\end{minipage}
	
	\caption{\label{fig:limitation} Example of imperfect boundary in the base weight maps. Insets are magnified $\times 4$. Best viewed on screen.}
\end{figure}
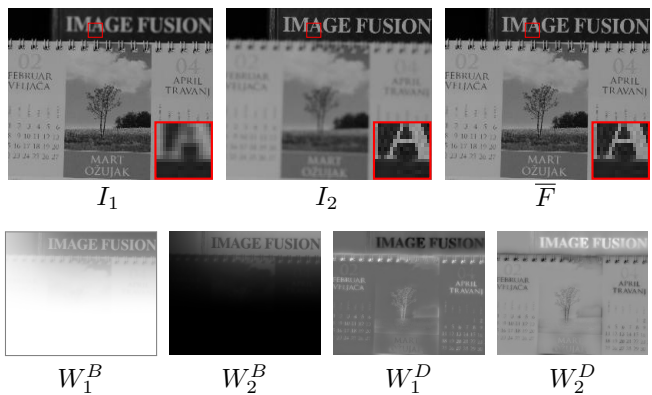

\subsection{Limitations}
In the multi-focus fusion task, we do not rely on any boundary detection method to refine the weight maps. The guided filter provides a good estimate of the boundaries between the multi-focus images, but does not always find the perfect boundaries. Fig.~\ref{fig:limitation} shows an example where the base weight maps do not accurately represent the focus regions, leading to imperfectly focused fusion results. LIU-M~\cite{liu2017multi} proposes the use of morphological operators to reshape the boundary conditions and fill holes in the multi-focus weight maps generated by their neural networks. Such adjustments to the focus weight maps generated by our method could further improve its multi-focus fusion weights, and consequently the resulting all-in-focus images.

%viff: 102
%qmi: 166
%qg: 1193
%qy: 215
%qc: 122
%qp: 65

%auto-ignore
\section{Conclusion}
We present a novel image fusion algorithm based on saliency and pre-trained neural networks. We first decompose source images into a base and a detail layer. Then, visual saliency maps and deep feature maps are used to compute base and detail fusion weights, respectively. Unlike typical neural network based techniques, our method requires no prior training on the image modalities and generalizes well to cover different fusion tasks. We demonstrate the robustness of our technique to the choices of decomposition filter, network architecture and feature depth.

Due to its robustness, we can configure our method to generate extremely fast and high-quality images, obtaining state-of-the-art results. We also demonstrate its applicability to diverse image fusion tasks, namely thermal, medical, and multi-focus fusion. Additionally, we show that our method can be extended to any number of input images. In conclusion, our method is a lightweight and high-quality technique with promising applications in real time systems and on low-energy hardware.

\section*{Acknowledgment}
We thankfully acknowledge the support of the Hasler Foundation (grant no. 16076, S.A.V.E.) for this work.

\ifCLASSOPTIONcaptionsoff
  \newpage
\fi



%\bibliographystyle{IEEEtran}
\bibliography{bare_jrnl}
\bibliographystyle{IEEEtran}

\newpage
\begin{IEEEbiography}[{\includegraphics[width=1in,height=1.25in,clip,keepaspectratio]{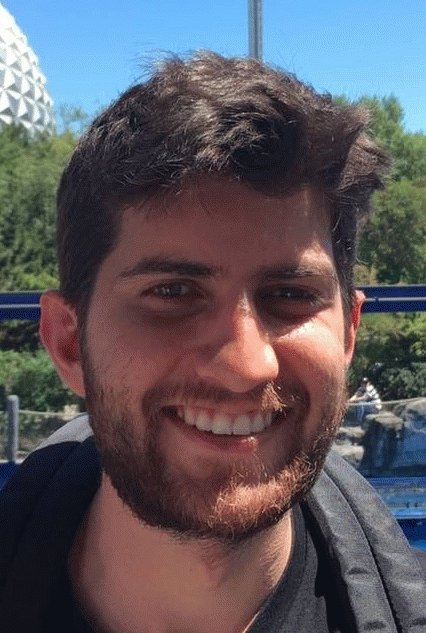}}]{Fayez Lahoud}
received his B.E. in Computer and Communication Engineering and his minor in Mathematics from the American University of Beirut, Lebanon. He received his M.S. in Computer Science at \'Ecole Polytechnique F\'ed\'erale de Lausanne where he is currently pursuing his Ph.D. at the Image and Visual Representation Laboratory. His work focuses on the development of computational and visual tools to help firefighters accomplish their tasks more efficiently.\\
\end{IEEEbiography}
\begin{IEEEbiography}[{\includegraphics[width=1in,height=1.25in,clip,keepaspectratio]{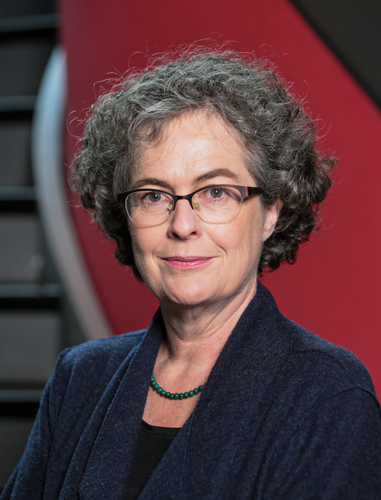}}]{Sabine S\"{u}sstrunk}
leads the Images and Visual Representation Lab (IVRL) at EPFL, Switzerland. Her research areas are in computational photography, color computer vision and color image processing, image  quality, and computational aesthetics. She has published over 150  scientific papers, of which 7 have received best paper/demos awards, and  holds 10 patents. She received the IS\&T/SPIE 2013 Electronic Imaging  Scientist of the Year Award and IS\&T’s 2018 Raymond C. Bowman Award. She is a Fellow of IEEE and IS\&T.
\end{IEEEbiography}
\vfill

%\newpage


%\vfill


\end{document}